\documentclass[10pt,twocolumn,letterpaper]{article}

\usepackage[pagenumbers]{cvpr}      

\usepackage[table]{xcolor}
\usepackage{pifont}
\usepackage[utf8]{inputenc}
\usepackage[normalem]{ulem} 
\usepackage{tikz}

\newcommand{\coluline}[3][1pt]{%
  \tikz[baseline=(text.base)]{
    \node[inner sep=0pt, outer sep=1pt, anchor=base] (text) {#3};
    \draw[#2, line width=#1] 
      ([yshift=-0.5pt]text.south west) -- 
      ([yshift=-0.5pt]text.south east);
  }%
}

\definecolor{cvprblue}{rgb}{0.21,0.49,0.74}
\usepackage[pagebackref,breaklinks,colorlinks,allcolors=cvprblue]{hyperref}
\usepackage{multirow}
\usepackage{cuted}
\usepackage[dvipsnames]{xcolor}
\def\paperID{XXXXX} 
\def\confName{CVPR}
\def\confYear{2026}
\usepackage{algorithm}
\usepackage{algpseudocode}
\usepackage{float}
\usepackage{tikz}
\usepackage{wrapfig}
\newcommand{\mygreenbox}[1]{\cellcolor{HighlightGreen}$\mathbf{#1}$}
\newcommand{\mygreenboxComp}[1]{\cellcolor{HighlightGreen}#1}
\definecolor{HighlightGreen}{HTML}{E6FFE6} 

\usepackage{fontawesome5}
\usepackage{xcolor}
\usepackage[table]{xcolor} 
\usepackage[most]{tcolorbox}

\newtcolorbox{promptbox}{
  colback=gray!10,
  colframe=gray!40,
  boxrule=0.5pt,
  arc=2pt,
  left=6pt,
  right=6pt,
  top=6pt,
  bottom=6pt
}

\title{ iSHIFT\ 
\hspace{-0.6em}\raisebox{-1.5em}{\includegraphics[height=3.5em]{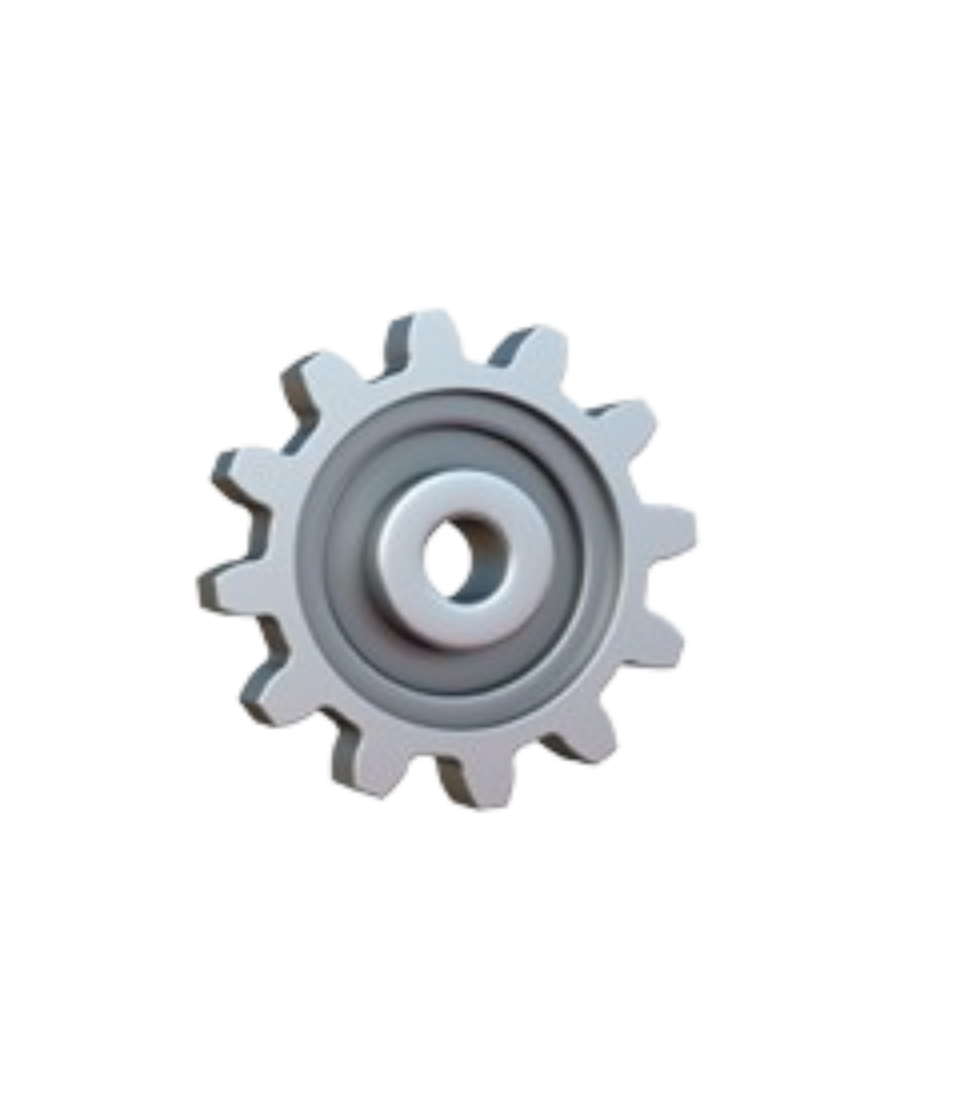}}\hspace{-1em}
:\! Lightweight Slow-Fast GUI Agent with Adaptive Perception\vspace{-20pt}}
\author{Sarthak Mehrotra\\
Indian Institute of Technology, Bombay\\
{\tt\small sarthak2002.mehrotra@gmail.com}
\and
Sairam VC Rebbapragada\\
Indian Institute of Technology, Hyderabad\\
{\tt\small ai20resch13001@iith.ac.in}
\and
Mani Hemanth Reddy Bonthu\\
Indian Institute of Technology, Hyderabad\\
{\tt\small 
cs22btech11013@iith.ac.in}
\and
Vineeth N Balasubramanian\\
Indian Institute of Technology, Hyderabad\\
{\tt\small vineethnb@cse.iith.ac.in}
}

\begin{document}
\maketitle

\begin{strip}
\centering
\vspace{-6pt}
\includegraphics[width=0.95\linewidth]{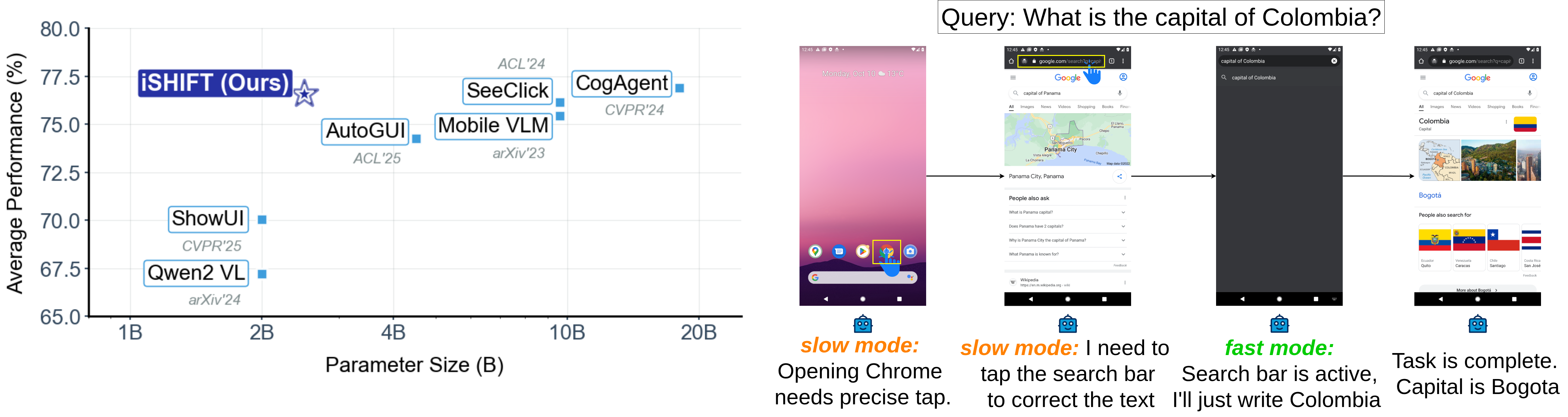}
\captionof{figure}{\textbf{Left}: Compared to recent state-of-the-art methods on Android In The Wild Benchmark \cite{NEURIPS2023_bbbb6308}, our approach achieves a competitive performance despite smaller size. \textbf{Right}: iSHIFT achieves this by adaptively switching between slow and fast modes based on task needs.}
\vspace{-6pt}
\label{fig:teaser}
\end{strip}

\begin{abstract}
Multimodal Large Language Models (MLLMs) show strong potential for interpreting and interacting with complex, pixel-rich Graphical User Interface (GUI) environments. However, building agents that are both efficient for high-level tasks and precise for fine-grained interactions remains challenging. GUI agents must perform routine actions efficiently while also handling tasks that demand exact visual grounding, yet existing approaches struggle when accuracy depends on identifying specific interface elements. These MLLMs also remain large and cannot adapt their reasoning depth to the task at hand. In this work, we introduce \textbf{iSHIFT}: \textbf{I}mplicit \textbf{S}low–fast \textbf{H}ybrid \textbf{I}nference with \textbf{F}lexible \textbf{T}okens, a lightweight agent that integrates latent thinking (implicit chain-of-thought) with a perception control module. \textbf{iSHIFT} enables an MLLM to switch between a slow mode, which leverages detailed visual grounding for high precision and a fast mode that uses global cues for efficiency. Special perception tokens guide attention to relevant screen regions, allowing the model to decide both how to reason and where to focus. Despite its compact \textbf{2.5B} size, \textbf{iSHIFT} matches state-of-the-art performance on multiple benchmark datasets. 
\end{abstract}    
\section{Introduction}
\label{sec:introduction}


Agents capable of understanding and operating Graphical User Interfaces (GUIs) are emerging as a pragmatic approach for automating everyday digital tasks. Acting as virtual intermediaries, such agents can mitigate user effort, orchestrate complex workflows, and enhance accessibility across mobile, desktop, and web platforms \cite{liu2024visualwebbench,rawles2025androidworld,NEURIPS2023_bbbb6308,lu2024gui}. Initial research efforts converged on language-driven pipelines contingent upon \emph{structural oracles} such as HTML, Document Object Model (DOM), or accessibility trees (AXTree) \cite{NEURIPS2023_5950bf29,zheng2024seeact}. While effective under ideal conditions, these methodologies suffer from poor generalization, as they are often tied to platform-specific inconsistencies and require complex preprocessing heuristics \cite{nguyen2025gui, qin2025ui}. Moreover, the frequent unavailability or the need for special permissions to access this structural information at runtime curtails their practical deployment.


These constraints prompted a shift towards vision-based and vision-language agents that operate directly upon screenshot-derived pixel data, mirroring human perceptual input, to generate grounded actions \cite{cheng2024seeclick,lu2024omniparserpurevisionbased,yan2023gpt}. Nevertheless, engineering reliable agents from raw pixels remains a challenge. GUIs contain fine-grained text, icons, and interactive elements within highly variable layouts, demanding both meticulous perception and robust contextual reasoning. Many existing approaches mitigate this by employing resource-intensive perception stacks, incorporating Optical Character Recognition (OCR), segmentation, or specialized backbones to enhance accuracy. This strategy, however, imposes a significant computational burden, as heavy perception modules remain constantly active, incurring undue latency and cost even for trivial actions \cite{Hong_2024_CVPR,author2024vzen,you2024ferretuigroundedmobileui,shen2024falconuiunderstandingguifollowing}.


Additionally, a central limitation pervading the GUI agents is their absence of \emph{adaptive} resource allocation. While slow-fast paradigms are studied in general-purpose visual agents \cite{sun2025visual, christakopoulou2024agents}, their application to GUI navigation remains nascent. Consequently, most GUI agents process all actions uniformly, disregarding the natural dichotomy within UI interactions between \emph{fast actions} (e.g., swiping) executable with global context, and \emph{slow actions} (e.g., finding a small button) that necessitate deliberate reasoning or enhanced visual precision. This uniform treatment compromises the balance between efficiency and reliability: models either expend excessive computation on simple tasks or fail to allocate sufficient reasoning for difficult ones. A recent work \cite{tang2025thinktwiceclickonce} attempts to make GUI automation adaptive, however, it relies on generating \emph{explicit intermediate representations} (e.g., icon captions) to improve grounding. For complex tasks, its slow path activates a text-generation loop, forcing the model to generate interface summaries and visual analyses before it can act. This approach incurs generative latency and couples grounding accuracy to the quality of the generated text. This multi-stage process trades one set of problems for another and highlights the need for a more integrated solution.

To address these limitations, we introduce \textbf{iSHIFT} (\textbf{I}mplicit \textbf{S}low–fast \textbf{H}ybrid \textbf{I}nference with \textbf{F}lexible \textbf{T}okens), a \(\sim2.5\)B-parameter multimodal GUI agent that natively integrates both compute and perception adaptivity within a unified model. iSHIFT is founded on the principle that the model itself should \emph{implicitly} determine when to engage in deeper, \emph{slow} reasoning. It achieves this through two complementary mechanisms: \textbf{latent thinking tokens}, which facilitate non-verbal internal deliberation, and a \textbf{perception token}, which on-demand activates a lightweight, localized visual perception module. This co-design obviates the need for external controllers, tightly couples reasoning depth with selective visual acuity, and achieves both efficiency on simple interactions and precision on complex ones. As depicted in Figure~\ref{fig:teaser}, our framework enables a compact agent to achieve state-of-the-art performance-to-size trade-off (Fig.~\ref{fig:teaser} left), by adaptively switching between \emph{fast} and \emph{slow} reasoning modes during action episodes (Fig.~\ref{fig:teaser} right).


Our main contributions are as follows:
\begin{enumerate}
    \item We introduce \textbf{iSHIFT}, a unified multimodal agent that integrates intrinsic slow-fast control. It leverages latent thinking tokens to \textbf{implicitly} and \textbf{on-demand} trigger a lightweight visual perception module, thereby eliminating external controllers and intermediate representations.
    
    \item We propose a novel training strategy for instilling adaptive control. By programmatically annotating training data to conditionally include perception tokens for high-precision tasks, the model learns to correlate task complexity with its implicit slow-fast mechanisms.
    
    \item We demonstrate that our agent achieves state-of-the-art performance-to-compute trade-off. Across multiple benchmarks, iSHIFT either surpasses or matches substantially larger models in performance while utilizing much fewer parameters.
\end{enumerate}
\section{Related Work}
\label{sec:related_works}

\noindent \textbf{Vision-Based GUI Agent Architectures.}
Early GUI automation efforts often relied on structured interface representations like Document Object Model (DOM) trees or platform-specific accessibility data \cite{ijcai2021p235, NEURIPS2023_5950bf29}. While effective in constrained environments like web pages, these methods lack generality, as such structural information is often unavailable or inconsistent across arbitrary desktop and mobile applications. The rise of powerful vision-language models (VLMs) has enabled agents that operate directly from pixel-level observations, closely mirroring human perception \cite{yan2023gpt, NEURIPS2023_6c52a8a4, cheng2024seeclick, chu2023mobilevlm, Hong_2024_CVPR, author2024vzen, zhang2025tongui, Huang_2025_CVPR, showui}. To improve UI understanding, \cite{cheng2024seeclick} introduces grounding-based pretraining, while subsequent work highlights a core challenge: GUIs are high-resolution and densely populated with small but crucial elements (e.g., icons, text fields) that are easily lost under naive downsampling. To preserve fine-grained details, several methods employ high-resolution encoders \cite{Hong_2024_CVPR, author2024vzen}, improving grounding at the cost of significant computational overhead. \cite{Huang_2025_CVPR} addresses locational ambiguity in dynamic high-resolution interfaces through Universal Block Parsing (UBP), enhancing GUI element localization. Meanwhile, \cite{showui} mitigates token inefficiency by introducing UI-guided visual token selection, grouping visually similar regions to reduce the processing burden.


\noindent \textbf{Slow-Fast Approaches.}
To mitigate the trade-off between efficiency and fidelity, a \emph{slow-fast} paradigm has emerged. These methods explicitly acknowledge that not all tasks require the same level of computation. However, existing implementations introduce new architectural complexities that undermine their benefits. For example, \cite{sun2025visual} proposes a system that arbitrates between a \emph{fast} and \emph{slow} path, but it relies on an external controller to make this decision. This disjointed design introduces modular complexity, adds latency, and creates a new potential point of failure. Similarly, GUI-specific work like \cite{tang2025thinktwiceclickonce} improves precision by generating explicit intermediate representations (e.g., icon captions) as a prerequisite slow grounding step. This approach, while effective, exacerbates system complexity and inference time by forcing the model to generate verbose text before it can act. Both methods validate the need for adaptive computation but highlight a research gap: the lack of a unified and implicit control mechanism. 

\noindent \textbf{Thought-Based GUI Agents.}
Another line of research enhances agent reliability by incorporating explicit cognitive loops, often inspired by Chain-of-Thought (CoT) reasoning \cite{NEURIPS2022_9d560961}. Agents such as \cite{wang2024mobile} adopt prompt frameworks like ReAct \cite{yao2022react}, interleaving reasoning steps with actions to form explicit plans and reflect on intermediate observations. \cite{wu2025osatlas} strengthens action prediction by generating thoughts that analyze instructions, screenshots, and interaction history. Similarly, \cite{xu2025aguvis} leverages structured reasoning to decompose complex tasks, adapt to new scenarios, and plan actions more effectively. However, a key limitation of these approaches is their dependence on explicit natural language generation. This verbosity incurs substantial inference latency, which is problematic for interactive GUI tasks where rapid responses are essential.
\vspace{-6pt}

\definecolor{LPT}{HTML}{9999FF}
\definecolor{PM}{HTML}{00CCCC}
\definecolor{LTT}{HTML}{F5D2D2}

\begin{figure*}[t]
    \centering
    \includegraphics[width=0.97\linewidth]{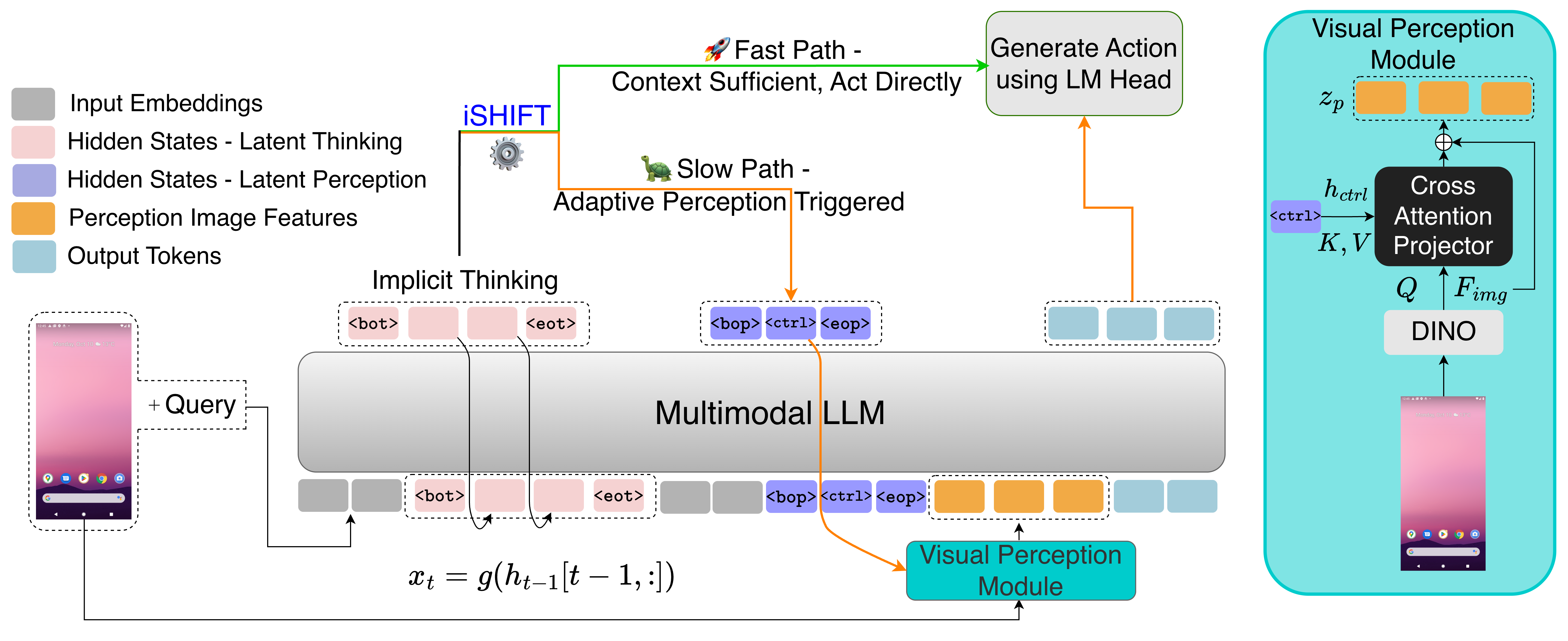}
    \vspace{-10pt}
    \caption{\textbf{Overview of our proposed method}. The MLLM takes a screenshot and task query as input and begins with the {\color{Green}Fast Path}. It utilizes the {\color{CarnationPink}Latent Thinking Tokens} (\texttt{<bot>}...\texttt{<eot>}) for implicit thinking to assess if the context is sufficient for direct action. If not, it switches to the {\color{orange}Slow Path} by generating {\color{LPT}Latent Perception Tokens} (\texttt{<bop>}, \texttt{<ctrl>}, \texttt{<eop>}). This invokes the {\color{PM}Visual Perception Module} which extract localized image features ($z_p$) for precise and grounded action generation.}
    \vspace{-7pt}
    \label{fig:architechture}
\end{figure*}

\section{iSHIFT: Methodology}
\label{sec:methodology}
\vspace{-3pt}
We now describe our iSHIFT methodology, which is centered on the core principles of implicit thinking and adaptive resource allocation. We begin by explaining the primary contribution: the iSHIFT Slow-Fast Approach (\S  \ref{subsec:slow_fast_approach}), which forms the operational core of our agent. This mechanism allows the model to implicitly shift between a computationally efficient \emph{{\color{Green}Fast Path}} for simple actions and a high-precision \emph{{\color{orange}Slow Path}} for complex ones. 
We then detail the two key Architectural Components that enable this adaptability: the {\color{CarnationPink}Latent Thinking Tokens} for general deliberation, and the {\color{PM}Visual Perception Module} which provides the localized features for the slow path.

\vspace{-2pt}
\subsection{The iSHIFT Slow-Fast Approach}
\label{subsec:slow_fast_approach}
\vspace{-3pt}
The central innovation of iSHIFT is an adaptive processing mechanism that addresses the dichotomy in UI tasks: simple actions (e.g., swipe left) can be executed with global context, while complex actions (e.g., tap tiny icon) demand precise, localized visual processing. Our model addresses this by following a sequential process detailed in Algorithm \ref{alg:adaptive_processing}. Our process uses a \emph{\color{Green} Fast Path} by default and allows the model to \emph{conditionally} activate a \emph{\color{orange}Slow Path} for added precision \emph{only when necessary}. We briefly describe our process herein, and then detail the components later in this section.

\vspace{5pt}
\noindent \textit{\textbf{Step 1: Implicit Thinking ({\color{Green}Fast Path}).}}
As shown in Figure \ref{fig:architechture}, our model always begins with the computationally efficient {\color{Green}Fast Path}, which processes a given user instruction and a UI screenshot. The model then performs implicit deliberation using the {\color{CarnationPink}\textbf{Latent Thinking Tokens}} (\texttt{<bot>}...\texttt{<eot>}), inspired by \cite{coconut}, that are detailed in \S \ref{subsec:arch_components}. This step serves as the model's internal assessment phase, where it evaluates the task's requirements based on the given context.

\vspace{5pt}
\noindent \textit{\textbf{Step 2: The Implicit Shift Decision.}}
Following the initial deliberation, the model makes a critical decision implicitly: ``Is the current information sufficient to act?'' This decision is not made by an external controller, but via tokenization on the model's next generative step.
\begin{itemize}[leftmargin=*]
    \item \textit{Stay on {\color{Green}Fast Path}:} If the deliberation concludes that the task is simple (e.g., a swipe), the model proceeds to generate the action directly, bypassing step 3.
    \item \textit{Activate {\color{orange}Slow Path}:} If the deliberation concludes that the task requires high precision (e.g., click on a small icon), the model instead generates special \textbf{{\color{LPT}Latent Perception Tokens}:} \texttt{<bop>}, \texttt{<ctrl>}, \texttt{<eop>}, similar to \cite{yu2025vpt}, to invoke the {\color{PM}Visual Perception Module} detailed in \S \ref{subsec:arch_components}.
\end{itemize}

\vspace{5pt}
\noindent \textit{\textbf{Step 3 (Optional): Conditional Activation ({\color{orange}Slow Path}).}}
The generation of the \texttt{<bop>} token provides an explicit signal that activates the {\color{orange}Slow Path}. This is not a restart but rather a conditional detour for feature enhancement. This token triggers the \textbf{\color{PM}Visual Perception Module}, which uses a DINO \cite{oquab2024dinov} encoder to extract localized, fine-grained features $z_p$ from the screen. These new features are then injected back into the model's sequence.

\vspace{5pt}
\noindent \textit{\textbf{Step 4: Final Action Generation.}}
The model, now armed with the initial global features and optionally the augmented, fine-grained features (from the {\color{orange}Slow Path}) generates the action.

\vspace{6pt}
\noindent \textbf{Training the Implicit Shift.}
This entire conditional process is learned end-to-end. We teach the model this ability through a dataset enhancement step, where we programmatically categorize the action space of our datasets based on their perception requirements to include our {\color{CarnationPink}latent thinking tokens} and {\color{LPT} latent perception tokens}, as required. This enables the model to learn a direct association between the contextual scenario and the correct execution path.
Any action that requires the prediction of precise coordinates (e.g., clicking on a UI element) is labeled as a \textit{Slow Action}. Prompt labels for such actions are annotated to include the {\color{LPT}latent perception tokens} (\texttt{<bop>}, \texttt{<ctrl>} and \texttt{<eop>}). Observing this token sequence prompts the model to engage the {\color{PM}Visual Perception Module} to acquire additional context required for more precise grounding. 
All other actions that can be inferred from global context (e.g., swiping) are labeled as \textit{Fast Actions}. They are modeled to follow the {\color{Green}Fast Path}, bypassing the {\color{PM}Visual Perception Module}. 
This mapping provides the necessary supervision to enable the model to learn the association between task types (inferred from the instruction and screen) and the capability to generate the \texttt{<bop>} token to activate the {\color{orange}Slow Path}.

In contrast to prior work which use an explicit external controller, our iSHIFT framework embeds this critical decision-making step implicitly within our unified MLLM architecture. As shown in Figure \ref{fig:architechture}, this implicit mechanism allows the model to perform internal, non-linguistic deliberation and decide whether sufficient context exists or if additional visual information is required. This eliminates reliance on external components, keeping the model's rationale embedded in its hidden states and providing it with context-awareness.

\begin{algorithm}[h]
\footnotesize
\caption{\textbf{Slow–Fast Adaptive Control in iSHIFT}}
\label{alg:adaptive_processing}
\begin{algorithmic}[1]
\Require Input context $C = \{I, T\}$, where $I$ is the UI screenshot and $T$ is the textual instruction
\Ensure Predicted action $A$
\vspace{1pt}

\State Initialize token sequence $\mathcal{S} \gets [\texttt{<bot>}]$
\State Initialize latent hidden state $h_0$
\While{not end of sequence}
    \State $h_t \gets \text{Transformer}(h_{t-1}, \mathcal{S}_t)$ \\
    \While{$t$ within $\texttt{<bot>}$ and $\texttt{<eot>}$}
        \State $x_t \gets g(h_{t-1}[t-1,:])$
        \State Model performs implicit reasoning over context

    \EndWhile
    \State Predict internal control decision $d_t \in \{\text{Fast}, \text{Slow}\}$
    \If{$d_t = \text{Slow}$}
        \State Generate latent control sequence: 
        \[
            \mathcal{S}_{ctrl} = \{\texttt{<bop>}, \texttt{<ctrl>}, \texttt{<eop>}\}
        \]
        \State Add latent control sequence $\mathcal{S} \gets S_{ctrl}$
        \State Extract control state $h_{ctrl} \gets h_t[\texttt{<ctrl>}]$
        \State $F_{img} \leftarrow \text{DINOEncoder}(I)$
        \State $z_p \leftarrow \text{CrossAttention}(F_{img}, h_{\text{ctrl}})$ 
        \State $S \leftarrow S + z_p$ 
    \EndIf
    
\EndWhile
\vspace{1pt}
\State Generate final action $A \gets \text{LM}(h_T)$
\end{algorithmic}
\end{algorithm}

\vspace{-10pt}
\subsection{Architecture Components}
\label{subsec:arch_components}
\vspace{-2pt}
The Slow-Fast approach is enabled by two key architectural additions to the Qwen2-VL 2B \cite{qwen2vl} base model.

\noindent \textbf{Latent Thinking Tokens.} We integrate a continuous latent thinking mechanism, to enable the model to perform introspection without generating explicit language tokens. As mentioned in \S \ref{subsec:slow_fast_approach}, this mechanism introduces special tokens, \texttt{<bot>} and \texttt{<eot>}, which indicate the beginning and end of a continuous thought block. Using these tokens, the model develops its reasoning in the latent space, without decoding intermediate language outputs.
\begin{table*}[t]
\centering
\small
\setlength{\tabcolsep}{6pt}
\renewcommand{\arraystretch}{1.2}
\resizebox{\textwidth}{!}{
\begin{tabular}{l l c c c c c c c c}
\hline
\textbf{Method} & \textbf{Base Model} & \textbf{Size} & \textbf{Vision Only} & \textbf{General} & \textbf{Install} & \textbf{Google Apps} & \textbf{Single} & \textbf{WebShop} & \textbf{Avg.}\\
\hline
\rowcolor{lightgray!40}\multicolumn{10}{c}{\textbf{Models with parameter count above 5B}} \\
\hline
LLaMa 2 \cite{touvron2023llama2openfoundation} \scriptsize \textcolor{gray!60}{(arxiv’23)}  & LLaMa 2 & 7B & \ding{55} & 28.56 & 35.18 & 30.99 & 27.35 & 19.92 & 28.40 \\
CoCo-Agent \cite{ma2024comprehensive} \scriptsize \textcolor{gray!60}{(arxiv’24)} & LLaMa 2 & 7B & \ding{55} & 69.92 & 80.60 & 75.76 & 88.81 & 74.02 & 77.82\\
MobileAgent \cite{wang2024mobile} \scriptsize \textcolor{gray!60}{(ICLR’24 w)} & LLaMa & 7B & \ding{55}& 55.80& 74.98& 63.95& 76.27 &63.61& 66.92 \\
\hline
GPT-4o \cite{openai2024gpt4ocard} \scriptsize \textcolor{gray!60}{(arxiv’24)} & -- & -- & \checkmark & 47.06 & 49.12 & 52.30 & 80.28 & 46.42 & 55.02 \\
OmniParser \cite{lu2024omniparserpurevisionbased} \scriptsize \textcolor{gray!60}{(arxiv’24)} & -- & -- & \checkmark & 48.30 & 57.80 & 51.60 & 77.40 & 52.90 & 57.60 \\
CogAgent \cite{Hong_2024_CVPR} \scriptsize \textcolor{gray!60}{(CVPR’24)} & Vicuna 7B & 18B & \checkmark & 65.38 & 78.86 & 74.95 & {\coluline[1pt]{black}{93.49}} & 71.73 & {\coluline[1pt]{black}{76.88}}\\
SeeClick \cite{cheng2024seeclick} \scriptsize \textcolor{gray!60}{(ACL’24)} & QwenVL & 9.6B & \checkmark & 67.60 & 79.60 & {\coluline[1pt]{black}{75.90}} & 84.60 & 73.10 & 76.16 \\
Mobile VLM \cite{chu2023mobilevlm} \scriptsize \textcolor{gray!60}{(EMNLP’24)} & QwenVL & 9.6B & \checkmark & 69.58 & 79.87 & 74.72 & 81.24 & 71.70 & 75.42 \\
SphAgent \cite{chai-etal-2025-amex} \scriptsize \textcolor{gray!60}{(ACL’25)} & InternLM & 7B & \checkmark & 68.20 & 80.50 & 73.30 & 85.40 & {\coluline[1pt]{black}{74.00}} & 76.28  \\ 
TongUI \cite{zhang2025tongui} \scriptsize \textcolor{gray!60}{(arxiv’25)} & Qwen2.5 VL & 7B & \checkmark & 67.6 & 76.3 & 73.5 & 79.9 & 69.1 & 73.3 \\

\hline
\rowcolor{lightgray!40}\multicolumn{10}{c}{\textbf{Models with parameter count below 5B}} \\
\hline
AutoGUI \cite{zhang-zhang-2024-look} \scriptsize \textcolor{gray!60}{(ACL’24)} & T5 + ViT & 4.5B & \checkmark & 68.24 & 76.89 & 71.37 & 84.58 & 70.26 & 74.27  \\
Qwen2 VL \cite{qwen2vl} \scriptsize \textcolor{gray!60}{(arxiv’24)} & Qwen2 VL & 2B & \checkmark & 61.40 & 71.80 & 62.60 & 73.70 & 66.70 & 67.24 \\
ShowUI \cite{showui} \scriptsize \textcolor{gray!60}{(CVPR’25)} & Qwen2 VL & 2B & \checkmark & 63.90 & 72.50 & 69.70 & 77.50 & 66.60 & 70.04\\
TongUI \cite{zhang2025tongui} \scriptsize \textcolor{gray!60}{(arxiv’25)} & Qwen2.5 VL & 3B & \checkmark & 65.6 & 75.10 & \textbf{74.50} & 77.00 & 65.8 & 71.6\\
\hline
\multirow{1}{*}{\textbf{iSHIFT}} & Qwen2 VL & 2.5B & \checkmark  & \mygreenbox{\coluline[1pt]{black}{\textbf{70.6}}} & \mygreenbox{\coluline[1pt]{black}{\textbf{80.82}}} & 71.64 & \textbf{86.03} & \mygreenboxComp{\textbf{72.60}} & \mygreenboxComp{\textbf{76.34}}\\
\hline
\end{tabular}
}
\vspace{-5pt}
\caption{Evaluation results across various subsets on the Android In The Wild \cite{NEURIPS2023_bbbb6308} Benchmark, measured by Action Matching Score. Best scores in the category $\mathbf{<5\text{B}}$ are \textbf{bolded}. Overall best scores are $\text{\coluline[1pt]{black}{underlined}}$. Cells in the iSHIFT row with a $\text{\colorbox{HighlightGreen}{Green fill}}$ demonstrate \textbf{SOTA performance parity}, meaning the score is within 2.0 points of the column’s maximum value.}
\vspace{-8pt}
\label{tab:aitw_results}
\end{table*}

This operation transforms the hidden state directly into the next input embedding, bypassing the vocabulary layer entirely. By operating fully within this latent space, the model achieves a deeper representational flow that captures implicit reasoning with significantly fewer steps compared to explicit Chain-of-Thought (CoT) generation. The resulting latent thinking process governs the model’s internal deliberation and is responsible for triggering the {\color{orange}Slow Path} when extended reasoning is required, ensuring resource-efficient yet contextually adaptive action generation.

The model learns to use these latent tokens via a specific training methodology. Initially, the model is trained on a smaller dataset where the thought token positions (\texttt{<bot>}...\texttt{<eot>}) are populated with explicit thoughts in natural language. This phase is crucial because it teaches the model the correct generation process and the expected content of its internal deliberations. Once this foundation is learned, the explicit thought tokens are replaced with the final latent thought tokens, giving the model a strong prior on the latent structure and the type of information it should process internally. Since thoughts are ultimately represented latently rather than in natural language, no additional thought annotations are required when scaling to larger datasets ($\approx50 \times$) improving overall scalability. More details about the training and implementation are presented in the \texttt{Supplementary}.



\vspace{3pt}
\noindent \textbf{Visual Perception Module.} The {\color{PM}Visual Perception Module} serves as the engine for the {\color{orange}Slow Path}. When activated by the \texttt{<bop>} token, the image is re-encoded into information-rich feature maps ($F_{img}$) using a DINO \cite{oquab2024dinov} vision encoder. These visual features then cross attend with the {\color{LPT}latent perception token's} hidden states ($h_{ctrl}$) through a cross-attention projector. This allows the model to selectively attend to and aggregate critical visual details.

The new context embedding $z_p$ is derived via cross-attention where the image features act as the Query ($Q$) and the {\color{LPT}latent perception token's} hidden state act as the Key and Value ($K,V$):
\begin{align*}
    Q = \text{Linear}_Q(F_{img}); &\hspace{6pt} K,V = \text{Linear}_{K,V}(h_{ctrl})
     \\ 
    z_p = \hspace{4pt} \text{softmax}&(\frac{QK^T}{\sqrt{d_k}})V + F_{img}
\end{align*}
where $d_k$ is the dimensionality of the query vector $Q$. 

The resulting context embedding $z_p$ is injected back into the MLLM sequence, providing the fine-grained control needed for precise grounding. Crucially, unlike methods such as CogAgent \cite{Hong_2024_CVPR} or V-Zen \cite{author2024vzen}, our {\color{PM}Visual Perception Module} is lighter and activated only when necessary. This allows iSHIFT to achieve superior precision without the large increase in parameter count that plagues high-resolution modules in existing models.





\section{Experiments}
\vspace{-4pt}

\noindent \textbf{Datasets, Baselines and Evaluation Metrics.}
We evaluate iSHIFT across diverse GUI benchmarks including Android In The Wild \cite{NEURIPS2023_bbbb6308} (AITW), Android Control \cite{li2024effectsdatascaleui}, GUIOdyssey \cite{lu2024gui}, and GUIAct \cite{chen-etal-2025-guicourse}. For comparison, we use state-of-the-art GUI agents such as \cite{cheng2024seeclick, Hong_2024_CVPR, showui, zhang2025tongui, li2024effectsdatascaleui} which represent various approaches to high-performance and high-resolution GUI processing. Our evaluation reports the Action Matching Score (AMS) for AITW (across General, Single, Install, Web Shopping, Google Apps subsets) and Android Control (High and Low subsets). For GUIOdyssey, we report the Success Rate, and for GUIAct (Web Single and Phone subsets), we provide both Type accuracy and Success Rate metrics. More details related to the above are mentioned in the \texttt{Supplementary}.

\noindent \textbf{Results and Observations.}
The iSHIFT model, built on a compact $2.5\text{B}$ parameter base, achieves an Average Action Matching Score (AMS) of $\mathbf{76.34}\%$, establishing it as the SOTA leader among models under $5\text{B}$ parameters and reaching near parity with the overall SOTA model, CogAgent \cite{Hong_2024_CVPR} ($18\text{B}$), with only a $0.54$ point difference ($76.34\%$ vs.\ $76.88\%$) despite being more than seven times smaller. Relative to its closest competitor in the small model category, AutoGUI \cite{zhan2023autoui} ($4.5\text{B}$), iSHIFT provides a gain of over $2.0$ points ($76.34\%$ vs.\ $74.27\%$) while remaining significantly smaller in size. Its adaptive processing capability is consistently reflected across individual subsets: iSHIFT sets new SOTA performance on the General and Install subsets with AMS scores of $\mathbf{70.6}\%$ and $\mathbf{80.82}\%$, respectively, and achieves the highest score in the Single subset within the $<5\text{B}$ category ($\mathbf{86.03}\%$), highlighting the effectiveness of its adaptive Perception Tokens for precise and efficient action grounding. On WebShop, it continues to lead its category with $\mathbf{72.60}\%$, demonstrating strong compositional reasoning in structured environments, and it performs competitively on Google Apps with $71.64\%$, maintaining robust generalization across diverse UI layouts and interaction types.

\begin{figure*}[t]
    \centering
    \includegraphics[width=0.95\linewidth]{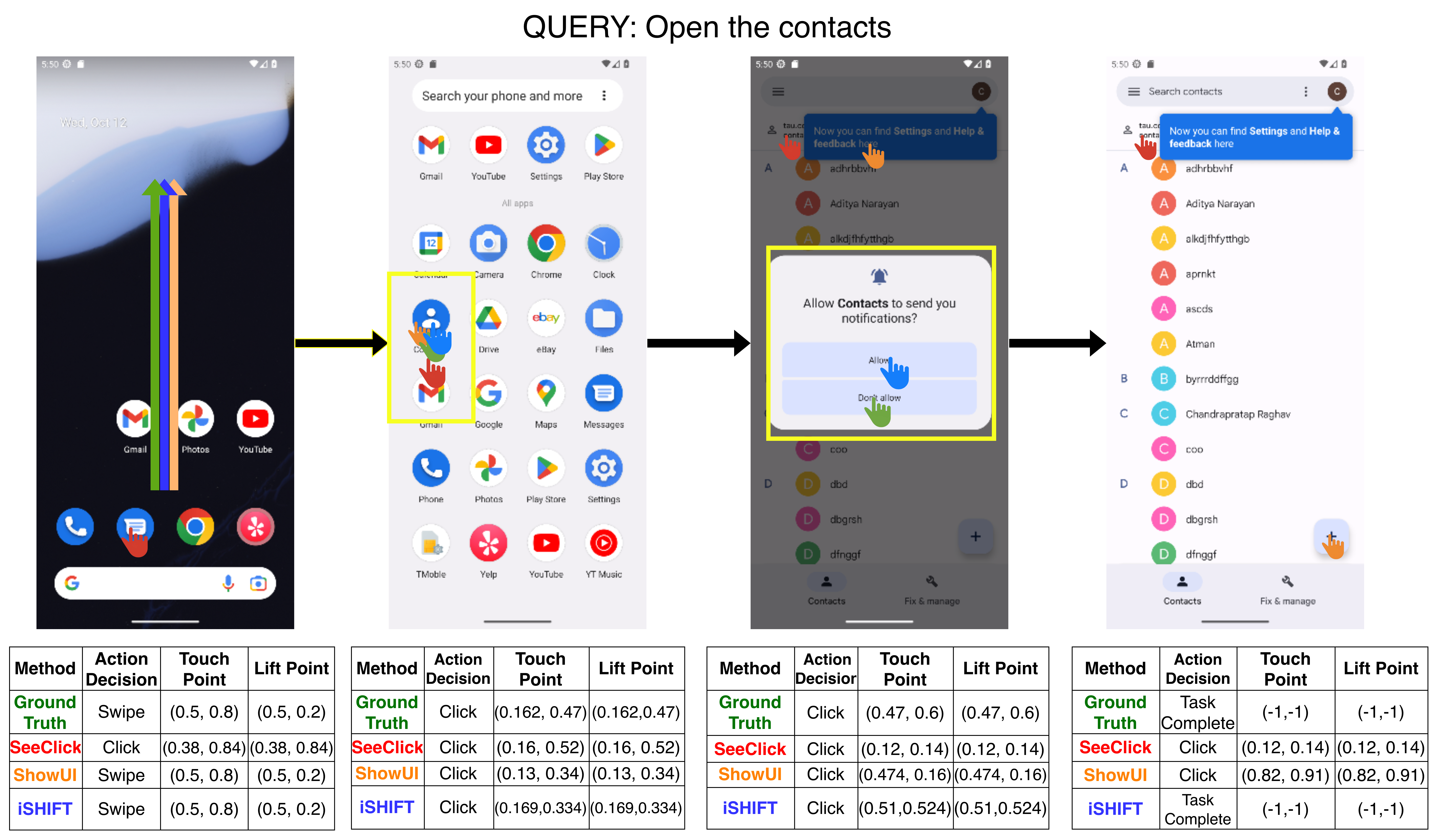}
    \vspace{-10pt}
    \caption{Qualitative comparison of GUI action sequences. The figure shows step-by-step actions taken SeeClick (red), ShowUI (orange), and iSHIFT (blue) in comparison to Ground Truth (green). Each colored hand or arrow represents the predicted interaction, including touch and lift points across interface states, and the ROI is highlighted with a yellow bounding box at each step. iSHIFT aligns closely with the correct action sequence, demonstrating superior spatial precision and decision accuracy compared to prior methods.}
    \vspace{-8pt}
    \label{fig:qualitative_result}
\end{figure*}
\begin{wrapfigure}[11]{r}{0.6\linewidth}
    \centering
    \vspace{-6pt}    
    \includegraphics[width=\linewidth]{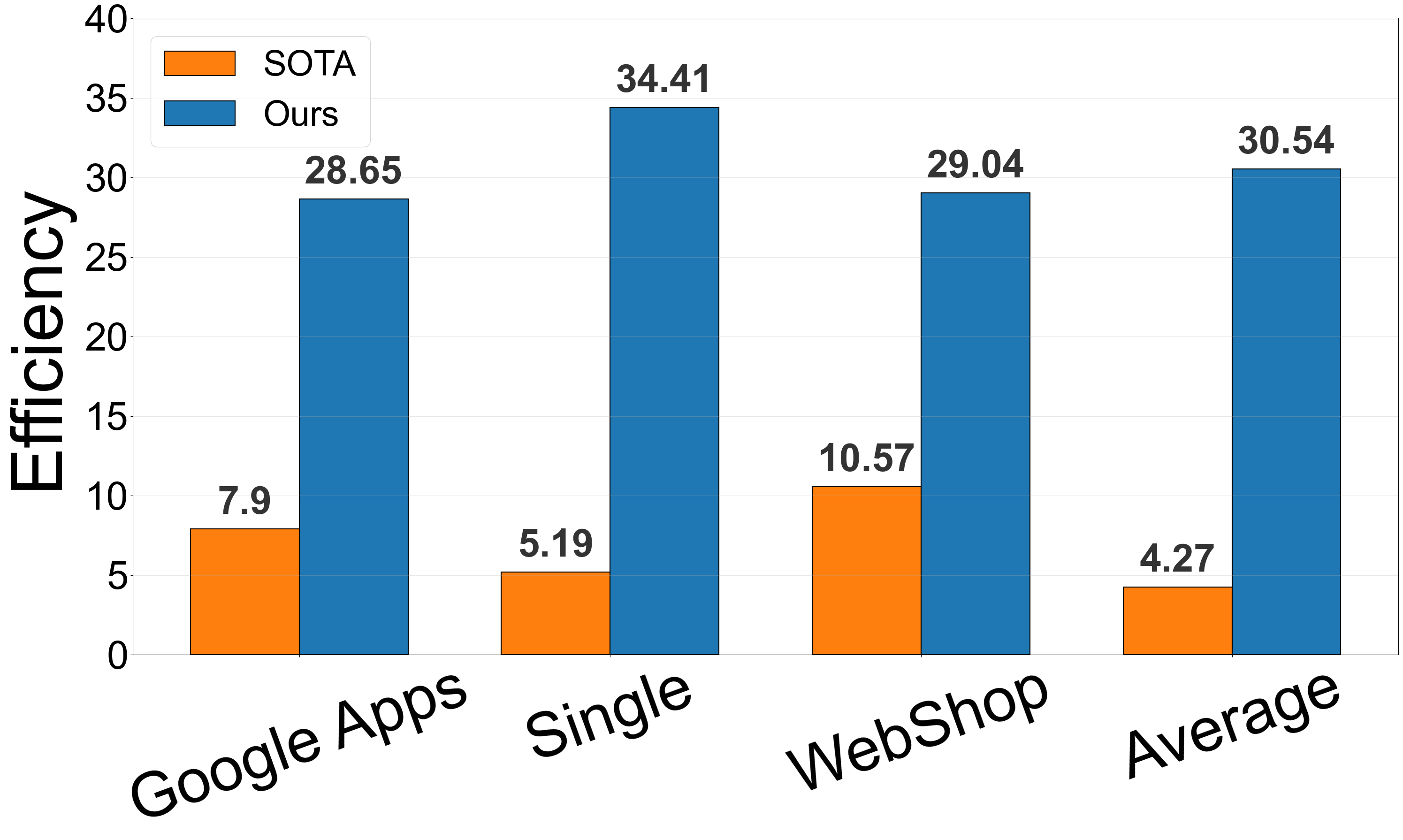}
    \vspace{-20pt}
    \caption{\footnotesize Efficiency comparison (Accuracy-to-Parameter ratio) of iSHIFT against subset wise state-of-the-art agents.}
    \label{fig:efficiency_metrics}
\end{wrapfigure}

The superior performance-to-size ratio of iSHIFT is clearly visualized in Figure \ref{fig:efficiency_metrics}, which plots the Efficiency Metric (Accuracy/Parameter Count) for iSHIFT compared to different SOTA models in the respective subsets. 
iSHIFT significantly outperforms all other SOTA models, achieving an efficiency of $\mathbf{30.54}$ on Average, which is more than five times that of the $18\text{B}$ CogAgent \cite{Hong_2024_CVPR} (which scores $\sim 4.27$ on Average). iSHIFT achieves a score of $\mathbf{28.65}$ in Google Apps, approximately four times that of the $9.6\text{B}$ SeeClick \cite{cheng2024seeclick} (which scores $\sim 7.9$). This provides concrete evidence that our adaptive token mechanism allows a highly compact $2.5\text{B}$ MLLM to achieve results comparable to substantially larger agents, setting a new standard for computational efficiency in resource-aware GUI agents.

The evaluation of $\text{iSHIFT}$ on cross-device benchmarks such as GUI Odyssey, summarized in Table \ref{tab:results_tab_2}, demonstrates strong generalization despite its compact $2.5\text{B}$ size, achieving the highest overall success rate of $\mathbf{73.97}\%$ across subsets and outperforming Aguvis ($63.8\%$), which is about four times larger. $\text{iSHIFT}$ also secures the top score in the Android Control (Low) subset with $\mathbf{87.7}\%$ and maintains competitive performance in the High-level precision subset with $65.6\%$, ranking just below $\text{OS Atlas}$ (4B), which achieves $67.54\%$ in the same category. As shown in Table \ref{tab:results_tab_3}, $\text{iSHIFT}$ continues to excel on the Web Single subset of GUIAct with a Type Accuracy of $\mathbf{93.83}\%$, surpassing $\text{GUICourse 3.1B}$ ($91.8\%$), while retaining a strong Success Rate of $66.38\%$. On the Phone subset, it attains $\mathbf{79.41}\%$ in Type Accuracy and $\mathbf{60.08}\%$ in Success Rate, outperforming the much larger $\text{GUICourse 9.6B}$ model, which records $73.0\%$ and $58.1\%$, respectively. These results highlight $\text{iSHIFT}$’s consistent high-fidelity performance across platforms while maintaining exceptional parameter efficiency.

\noindent \textbf{Qualitative Results} We present a qualitative comparison of GUI action sequences by evaluating iSHIFT, SeeClick and ShowUI against Ground Truth, in Figure \ref{fig:qualitative_result}. More qualitative results can be found in the \texttt{Supplementary}.

\begin{table}[h]
\centering
\footnotesize	
\setlength{\tabcolsep}{4pt}
\renewcommand{\arraystretch}{1.25}
\begin{tabular}{l c c c c}
\hline
\multirow{2}{*}{\textbf{Method}} & \multirow{2}{*}{\textbf{Size}} & \multicolumn{2}{c}{\textbf{Android Control}} & \multirow{2}{*}{\parbox{1cm}{\centering \textbf{GUI Odyssey}}} \\
& & High & Low & \\
\hline
\rowcolor{lightgray!40}\multicolumn{5}{c}{\textbf{Models with parameter count above 5B}}\\
\hline
GPT-4o \cite{openai2024gpt4ocard} \scriptsize \textcolor{gray!60}{(arxiv’24)} & -- & 28.39&21.17 & 5.36 \\
See Click \cite{cheng2024seeclick} \scriptsize \textcolor{gray!60}{(ACL’24)} & 9.6B & 59.1 & 75.0 &  53.92 \\
OS Atlas   \cite{wu2025osatlas} \scriptsize \textcolor{gray!60}{(ICLR’25)}  & 7B & \coluline[1pt]{black}{71.17} & 85.22 & 61.98 \\
TongUI \cite{zhang2025tongui} \scriptsize \textcolor{gray!60}{(arxiv’25)}  & 7B & 38.2 & 69.3 &  -- \\
Aguvis \cite{xu2025aguvis} \scriptsize \textcolor{gray!60}{(ICML’25)} & 7B & 61.5& 80.5 & 63.8 \\
\hline
\rowcolor{lightgray!40}\multicolumn{5}{c}{\textbf{Models with parameter count below 5B}}\\
\hline
Qwen2.5-VL \cite{Qwen2.5-VL} \scriptsize \textcolor{gray!60}{(arxiv’25)} & 3B & 58.79 & 41.22 & 27.31 \\
\multirow{1}{*}{OS Atlas}\cite{wu2025osatlas} \scriptsize \textcolor{gray!60}{(ICLR’25)} & 4B & \textbf{67.54} & 80.64 & 56.39 \\
\multirow{1}{*}{TongUI} \cite{zhang2025tongui} \scriptsize \textcolor{gray!60}{(arxiv’25)} & 3B & 32.0 & 52.0 &  -- \\
\hline
\textbf{iSHIFT} & 2.5B & 65.6 & \textbf{\coluline[1pt]{black}{87.7}} & \textbf{\coluline[1pt]{black}{73.97}}\\
\hline
\end{tabular}
\vspace{-5pt}
\caption{Evaluation results on Android Control \cite{li2024effectsdatascaleui} and GUI Odyssey \cite{lu2024gui} benchmarks. Best scores in the category $\mathbf{<5\text{B}}$ are \textbf{bolded}. Overall best scores are $\text{\coluline[1pt]{black}{underlined}}$.}
\label{tab:results_tab_2}
\vspace{-8pt}
\end{table}


\begin{table}[h]
\centering
\footnotesize	
\setlength{\tabcolsep}{4pt}
\renewcommand{\arraystretch}{1.25}
\begin{tabular}{l l l c c c}
\hline
\multirow{2}{*}{\textbf{Method}} & \multirow{2}{*}{\textbf{Size}} & 
\multicolumn{2}{c}{\textbf{Web Single}} & \multicolumn{2}{c}{\textbf{Phone}}\\
& & Type & SR  & Type & SR \\
\hline
\rowcolor{lightgray!40}\multicolumn{6}{c}{\textbf{Models with parameter count above 5B}}\\
\hline
GPT-4o \cite{openai2024gpt4ocard} \scriptsize \textcolor{gray!60}{(arxiv’24)} & -- & 77.09 & 41.84 & 46.9 &28.4 \\
GUICourse \cite{chen-etal-2025-guicourse} \scriptsize \textcolor{gray!60}{(ACL’25)} & 9.4B & 90.1  & 63.5 &72.1& 40.4  \\
GUICourse \cite{chen-etal-2025-guicourse} \scriptsize \textcolor{gray!60}{(ACL’25)} & 9.6B & 90.9  & 66.7 &73& 58.1 \\
OS-Atlas \cite{wu2025osatlas} \scriptsize \textcolor{gray!60}{(ICLR’25)} & 7B & 86.95 & 57.02 & -- & --  \\
\hline
\rowcolor{lightgray!40}\multicolumn{6}{c}{\textbf{Models with parameter count below 5B}}\\
\hline
Qwen2.5-VL \cite{Qwen2.5-VL} \scriptsize \textcolor{gray!60}{(arxiv’25)} & 3B & 56.1 &  55.61 & 60.8 &35.8 \\
GUICourse \cite{chen-etal-2025-guicourse} \scriptsize \textcolor{gray!60}{(ACL’25)} & 3.1B & 91.8  & \textbf{70.6} &71.7&  53.3 \\
OS-Atlas \cite{wu2025osatlas} \scriptsize \textcolor{gray!60}{(ICLR’25)} & 4B & 79.22  & 42.62 &--& --  \\
\hline
\textbf{iSHIFT} & 2.5B & \textbf{93.83} & 66.38 &\textbf{79.41}& \textbf{60.08} \\
\hline
\end{tabular}
\vspace{-5pt}
\caption{Evaluation results on Web Single and Phone subset of GUIAct \cite{chen-etal-2025-guicourse} benchmark. Overall best scores are \textbf{bolded}.}
\label{tab:results_tab_3}
\vspace{-10pt}
\end{table}


\begin{table}[t!]
\centering
\small
\setlength{\tabcolsep}{3pt}
\renewcommand{\arraystretch}{1.1}
\resizebox{\linewidth}{!}{
\begin{tabular}{l c c c c c c}
\hline
\textbf{Variation} & \textbf{VPM} & \textbf{LTT} & \textbf{Slow} & \textbf{Fast} & \textbf{AMS} \\
\hline
Base Model (Qwen2 VL 2B) & $\times$ & $\times$ & $\times$ & $\checkmark$ & 67.24\\
w/ Fast Path Only & $\times$ & $\checkmark$ & $\times$  & $\checkmark$ & 72.54\\
w/ VPM, Slow Path Only & $\checkmark$ & $\times$ & $\checkmark$ & $\times$  & 72.71\\
w/ VPM, Adaptive& $\checkmark$ & $\times$ & $\checkmark$ &$\checkmark$ &  75.48\\
w/ VPM + LTT, Slow Path Only & $\checkmark$ & $\checkmark$ & $\checkmark$ &$\times$ & 74.58\\

w/o Cross-Attention in VPM & $\checkmark$ & $\checkmark$ & $\checkmark$ & $\checkmark$ &73.40\\

\hline

iSHIFT (Ours) & $\checkmark$ & $\checkmark$ & $\checkmark$ & $\checkmark$ & \textbf{76.34}\\

\hline
\end{tabular}
}
\vspace{-5pt}
\caption{Ablation study on the AITW benchmark \cite{NEURIPS2023_bbbb6308} evaluating different components of iSHIFT. VPM denotes the Visual Perception Module and LTT refers to Latent Thinking Tokens.}
\vspace{-5pt}
\label{tab:ablation}
\end{table}

\vspace{-10pt}
\section{Analysis and Ablation Studies}
\label{sec:ablations}
\vspace{-4pt}
To better understand the effectiveness of our approach, we systematically evaluate different components of the proposed framework as presented in Table \ref{tab:ablation}. We evaluate how each module and design choice contributes to the overall performance on AITW \cite{NEURIPS2023_bbbb6308}. All other settings remain consistent with the full model unless stated otherwise.

\begin{table}[h!]
    \centering
    \footnotesize 
    \setlength{\tabcolsep}{5pt}
    \renewcommand{\arraystretch}{1.2}
    \begin{tabular}{l c c c c}
    \hline
    \multirow{2}{*}{\textbf{Method}} & \multicolumn{2}{c}
    {\textbf{AITW General}} & \multicolumn{2}{c}{\textbf{AITW Single}} \\
     & Runtime (ms) & AMS & Runtime (ms) & AMS \\
     \hline
     Fast&2093&66.64&2046&82.55 \\
    Slow&2331&68.2&	2323&85.33 \\
    \hline
    \textbf{OURS} &2229&70.6&2263&86.03\\
    \hline
     
    \end{tabular}
    \vspace{-5pt}
    \caption{Comparison of performance and latency between Adaptive iSHIFT and its fixed-path variants, showing that the adaptive version delivers higher accuracy with lower runtime.}
    \label{tab:slowfast_explanation}
    \vspace{-15pt}
\end{table}

\vspace{3pt}
\noindent \textbf{Impact of Individual Components.}
The standard Qwen2-VL-2B Baseline achieved an AMS of $67.24$ on AITW \cite{NEURIPS2023_bbbb6308} benchmark (Table~\ref{tab:ablation}). Adding Latent Thinking Tokens (LTT Only) boosts performance by $+5.30$ to $72.54$ (Row 2, Table~\ref{tab:ablation}), validating the intrinsic benefit of implicit deliberation for general task execution. Similarly, introducing the Visual Perception Module (VPM) to the baseline yields a comparable $+5.47$ gain, reaching $72.71$ (Row 3, Table~\ref{tab:ablation}).
The largest improvement arises from enabling path selection. When both slow and fast paths are available, a non-LTT-based implicit decision process (Row 4, Table~\ref{tab:ablation}) achieves $75.48$, a $+2.77$ increase over the slow-only variant (Row 3, Table~\ref{tab:ablation}), confirming the key role of path switching.
Finally, the Full iSHIFT model, combining LTTs with real-time adaptive control of the VPM, reaches the best AMS of $76.34$ (Row 7, Table~\ref{tab:ablation}). This LTT-driven decision-making contributes a further $+0.86$ beyond the non-LTT-based variant (Row 4, Table~\ref{tab:ablation}). Conversely, forcing iSHIFT to operate only on the slow path lowers performance to $74.58$ (Row 5, Table~\ref{tab:ablation}), showing that true adaptive switching adds $+1.76$ and is central to iSHIFT’s superiority.

\begin{figure}[h]
    \centering
    \includegraphics[width=0.8\linewidth]{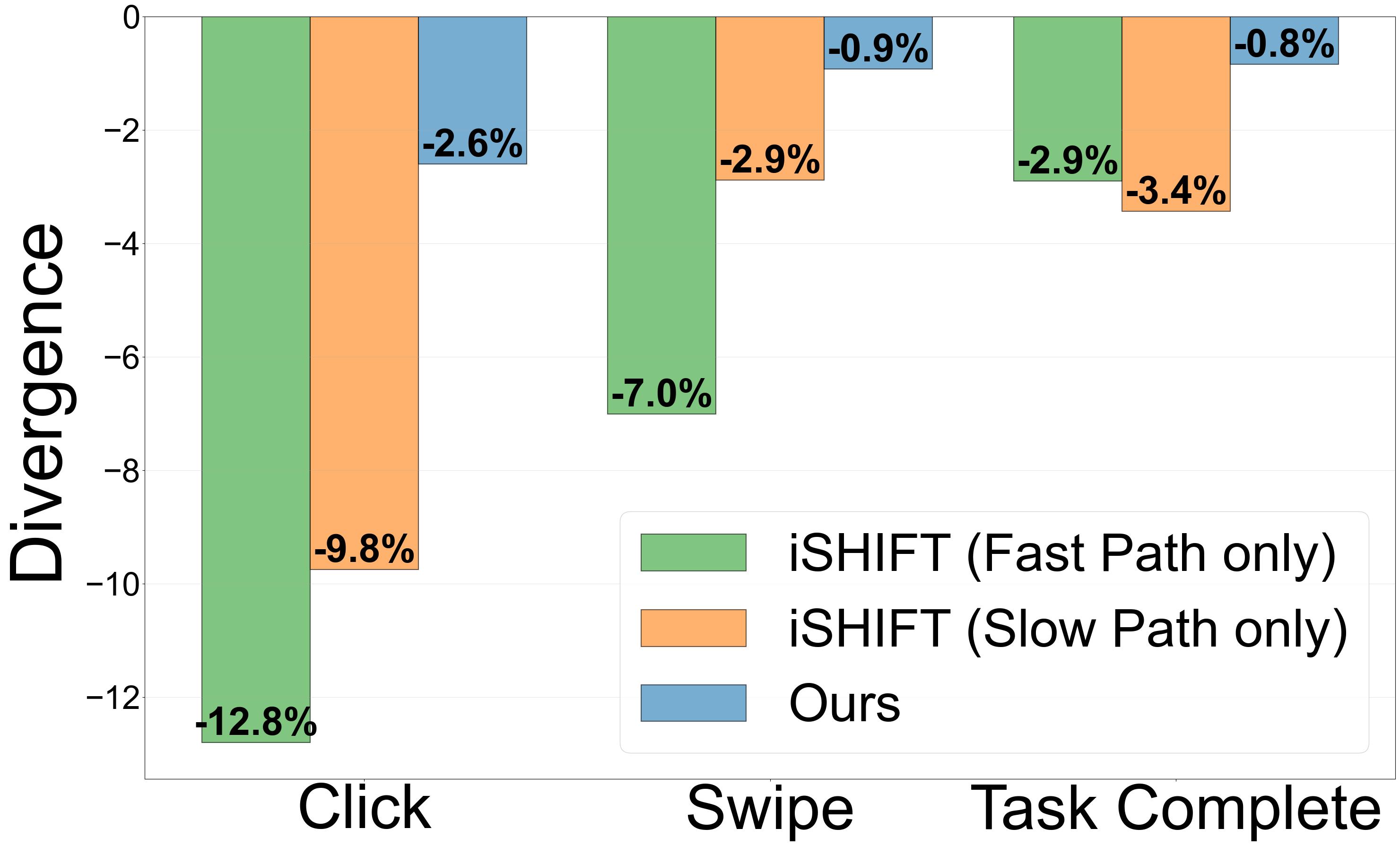}
    \vspace{-5pt}
    \caption{Divergence of distributions of predicted action from the Ground Truth across different iSHIFT variants (Smaller bar is better), This demonstrates the adaptive model's superior action fidelity and significantly lower error rates across all actions.}
    \label{fig:analysis_general}
    \vspace{-5pt}
\end{figure}

\begin{figure}[h]
    \centering
    \includegraphics[width=0.95\linewidth]{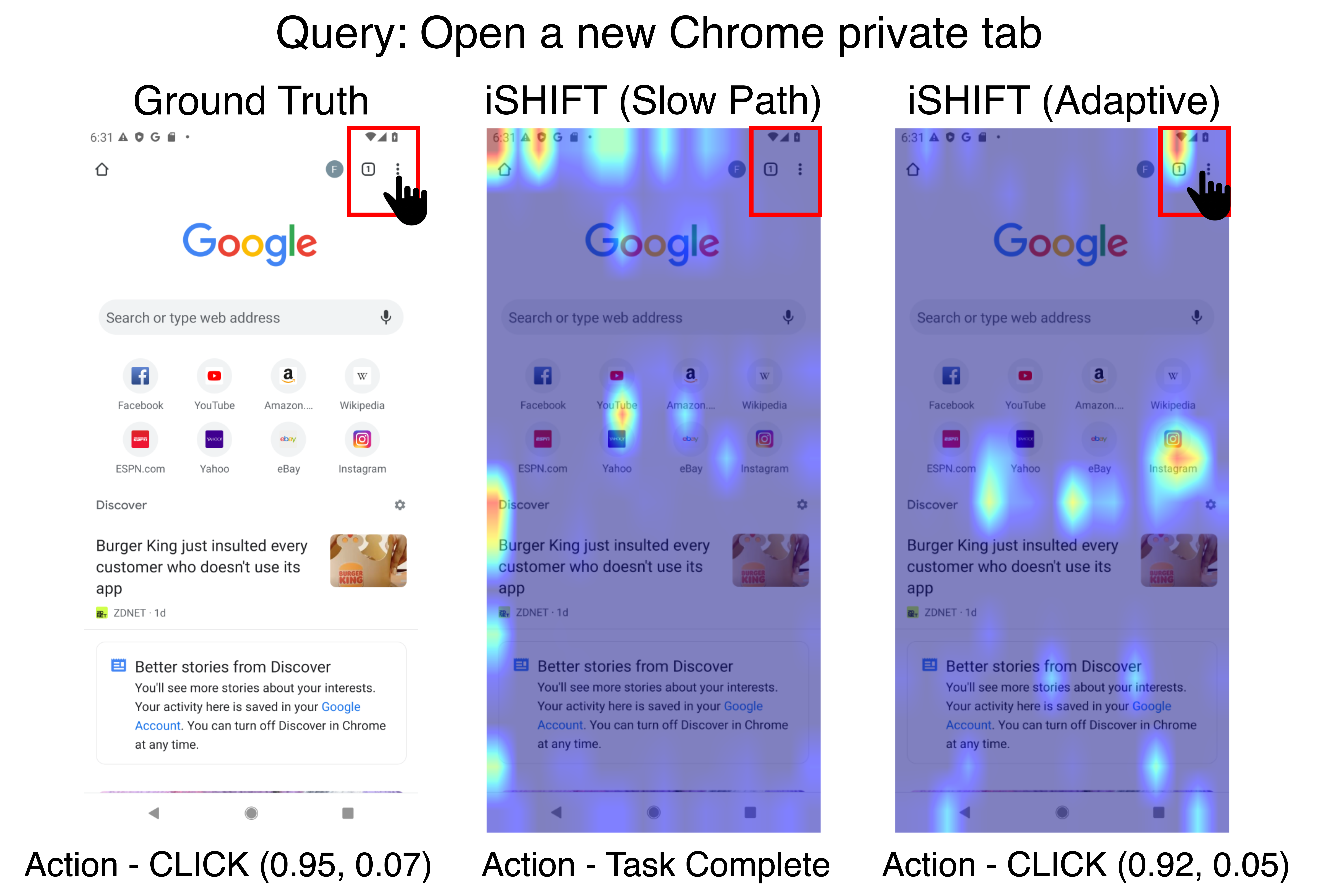}
    \vspace{-8pt}
    \caption{Attention maps from the cross-attention projector and the actions generated. This illustrates that mandatory localized image features in iSHIFT (Slow Path) produces diffused and irrelevant attention, leading to an incorrect action. In contrast, Adaptive iSHIFT (Ours) selectively attends to the correct target element, ensuring accurate action execution.}
    \label{fig:attention_maps}
    \vspace{-20pt}
\end{figure}

\noindent \textit{\textbf{Why is Adaptive Better than Always Slow?} }While the iSHIFT (Slow Path Only) variant achieves a strong AMS of $74.58$, it still lags behind the full adaptive model, suggesting that forced commitment to the slow path limits overall performance. As shown in Table \ref{tab:slowfast_explanation}, the adaptive iSHIFT model not only achieves higher AMS but also executes faster than the slow only variant, indicating that more computation time does not necessarily translate to better performance.

Figure \ref{fig:analysis_general} highlights that the adaptive model maintains a much smaller action type divergence from the Ground Truth across all key actions compared to the slow-only model.
For example, in complex click actions, its deviation is only $2.4\%$ compared to $8.9\%$ for the slow-only model. This discrepancy arises because the slow path variant processes detailed visual context indiscriminately for every action, even when such context is unnecessary. As illustrated in Figure \ref{fig:attention_maps}, this leads to diffused and misdirected attention, where the model struggles to distinguish different actions, often misclassifying actions such as predicting Task Complete instead of Click. The continual, unnecessary addition of visual context for simple actions (e.g., Task Complete) hinders decision clarity and leads to suboptimal behavior. In contrast, the adaptive iSHIFT selectively retrieves context only when needed, maintaining sharper attention on the true target element and ensuring the correct action is executed. Together, these results confirm that iSHIFT’s implicit adaptive control enhances both accuracy and efficiency, eliminating unnecessary over-processing on simpler tasks.

\vspace{3pt}
\noindent \textbf{Importance of Cross Attention Projector in VPM.} Removing cross-attention prevents the model from accessing localized visual features, causing the Slow Path to fall back to coarse DINO features and rely only on global context. This weakens precise grounding and leads to a reduced AMS of 73.40 (Row 6, Table~\ref{tab:ablation}).

\begin{wrapfigure}[11]{r}{0.6\linewidth}
\vspace{-5pt}
    \centering   
    \includegraphics[width=\linewidth]{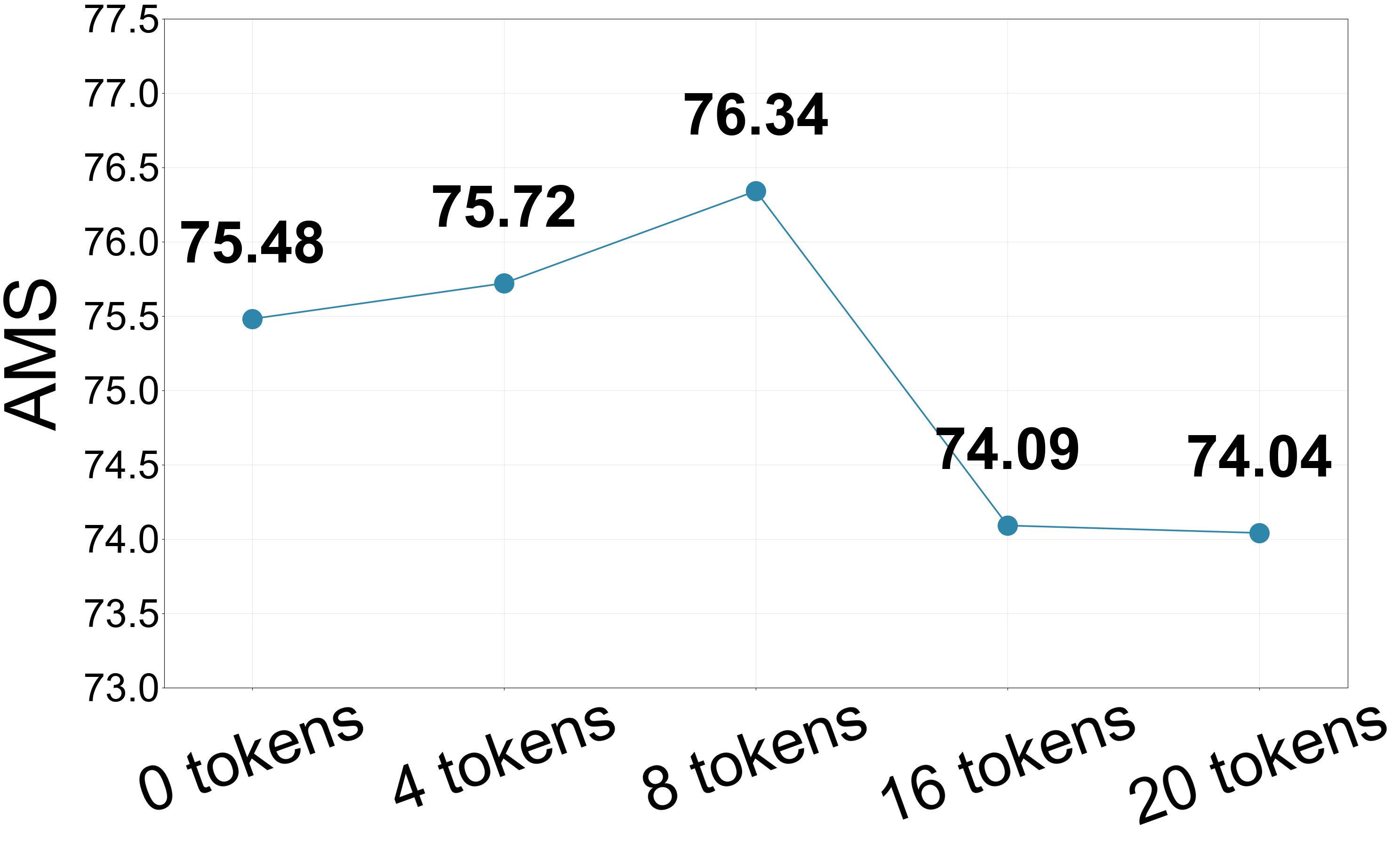}
    \vspace{-20pt}
    \caption{\footnotesize Impact of latent token count on average performance, showing optimal results around 8 tokens.}
    \label{fig:ablation_latent}
\end{wrapfigure}

\vspace{3pt}
\noindent \textbf{Effect of Number of Latent Tokens.} We conducted an ablation to determine the optimal number of latent tokens for effective implicit deliberation, since an overly long thought block increases computation without proportional gains. As shown in Figure \ref{fig:ablation_latent}, the baseline without latent tokens reached an AMS of $75.48$. Adding four latent tokens raised this to $75.72$, and using eight tokens yielded the best performance at $76.34$, providing sufficient reasoning depth with minimal overhead. Increasing the sequence further caused clear degradation, dropping to $74.09$ with sixteen tokens and $74.04$ with twenty. 
This non-linear trend shows that the length of the implicit thought block is a sensitive hyperparameter, and 
excessive deliberation can dilute contextual focus and hinder efficient sequence generation, ultimately reducing performance. We present more experiments related to Latent Thinking Tokens in the \texttt{Supplementary}.

\begin{wrapfigure}[11]{r}{0.6\linewidth}
\vspace{-5pt}
    \centering   
    \includegraphics[width=\linewidth]{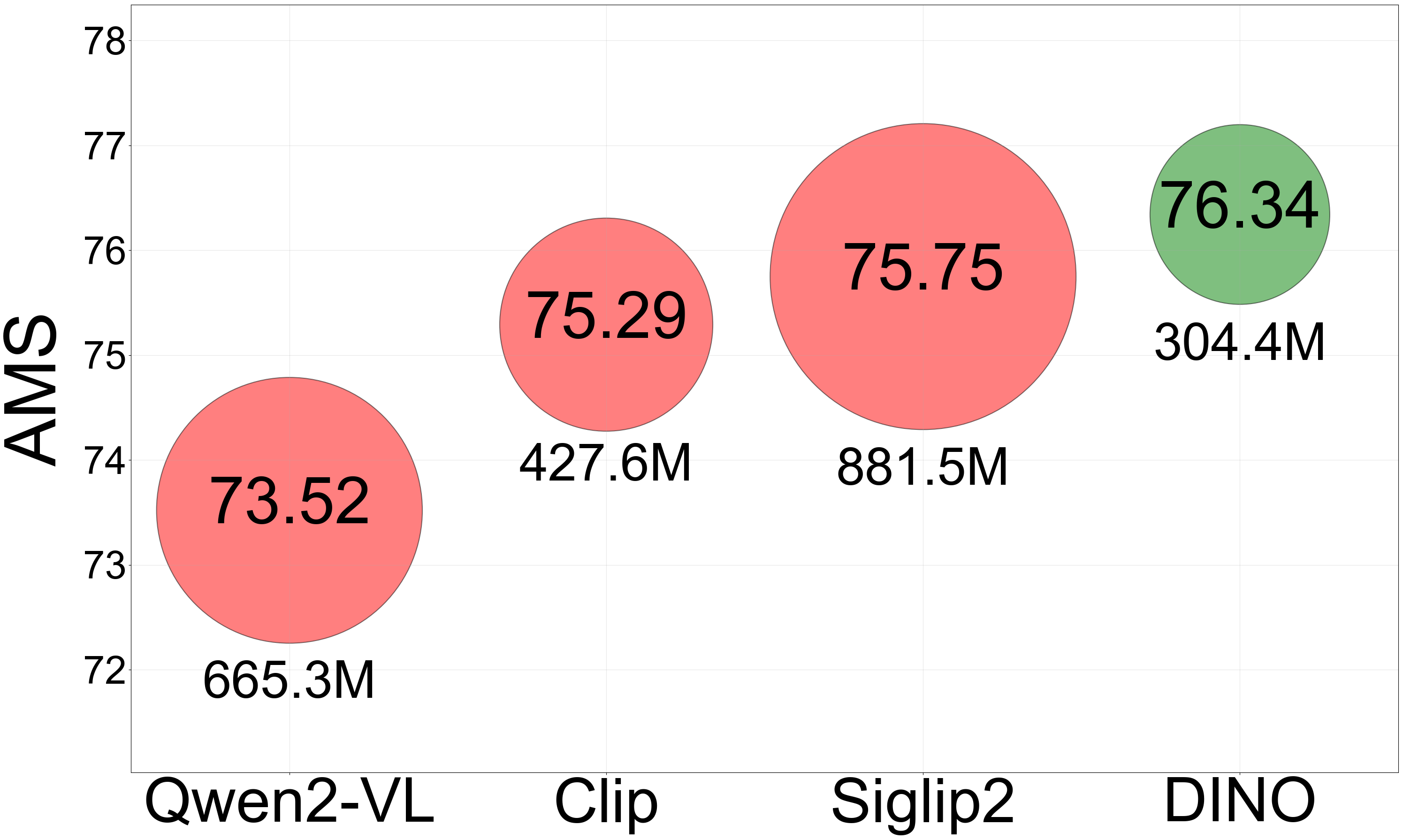}
    \vspace{-20pt}
    \caption{\footnotesize Performance of alternative visual encoders in the VPM. Circle size reflects encoder parameter count.} 
    \label{fig:ablations_pm}
\end{wrapfigure}

\vspace{3pt}
\noindent \textbf{Effect of Different Encoders in VPM.}
We conducted an ablation on the visual encoder used in the Visual Perception Module (VPM) to evaluate performance and efficiency, as shown in Figure \ref{fig:ablations_pm}. 
The standard Qwen2-VL Image Encoder achieved an AMS of $73.52$ with $665.3$M parameters. Replacing it with CLIP improved AMS to $75.29$ while reducing parameters to $427.6$M. Siglip-2 increased accuracy further to $75.75$ but required the largest parameter count at $881.5$M. Our final choice, the DINO encoder, reached the highest AMS of $76.34$ with only $304.4$M parameters. This confirms that DINO offers the best balance, providing high-quality visual features for adaptive grounding while remaining computationally lightweight for our $\text{iSHIFT}$ agent. We present more insight into the VPM in \texttt{Supplementary}.

\vspace{-5pt}
\section{Conclusion}
\vspace{-5pt}
We introduced iSHIFT, a compact multimodal GUI agent that integrates implicit slow–fast control with latent reasoning and selective perception. By combining latent thinking tokens with an on-demand Visual Perception Module, iSHIFT allocates resources adaptively, engaging in deeper visual grounding only when necessary. This design removes external controllers and maintains high precision across diverse interface layouts. Extensive experiments on AITW, GUI Odyssey, Android Control, and GUIAct show that iSHIFT matches or surpasses much larger models, establishing a new efficiency–accuracy trade-off for GUI agents. These results highlight implicit adaptive reasoning as a powerful approach for building capable and resource-aware multimodal agents.
{
    \small
    \bibliographystyle{ieeenat_fullname}
    \bibliography{main.bib}
}   
\onecolumn
\setcounter{section}{0}
\setcounter{figure}{0}
\setcounter{table}{0}
\renewcommand{\thesection}{S.\arabic{section}}
\renewcommand{\thetable}{S.\arabic{table}}
\renewcommand{\thefigure}{S.\arabic{figure}}
\renewcommand{\theequation}{S.\arabic{equation}}

{\centering
\section*{\Large  iSHIFT\ 
\hspace{-0.6em}\raisebox{-1.5em}{\includegraphics[height=3.5em]{image.png}}\hspace{-1em}
:\! Lightweight slow fast GUI Agent with Adaptive Perception \\Supplementary Material}}
\vspace{2pt}
\newcommand{\ToCEntry}[3]{%
  \ifcase#1
    \noindent\ref{#3}. #2\hspace{1.5em}\dotfill\hspace{1.5em}\pageref{#3} \\ 
  \or
    \noindent\hspace*{2em}\ref{#3}. #2\hspace{1.5em}\dotfill\hspace{1.5em}\pageref{#3} \\ 
  \else
    \noindent\ref{#3}. #2\hspace{1.5em}\dotfill\hspace{1.5em}\pageref{#3} \\ 
  \fi
}


\section*{Contents}
\ToCEntry{0}{Dataset Enhancement}{sec:dataset_enhancement}
\ToCEntry{0}{Reproducibility and Implementation details}{sec:implementation_details}
\ToCEntry{1}{Computation and Training Details}{subsec:compute}
\ToCEntry{1}{Hyperparameters}{subsec:hyperparams}
\ToCEntry{1}{Prompt Template}{subsec:prompt_template}
\ToCEntry{1}{Dataset Details}{subsec:dataset_details}
\ToCEntry{0}{Initilization of Latent Tokens}{sec:init}
\ToCEntry{0}{Chain of Thought v/s Latent Thinking}{sec:cot}
\ToCEntry{0}{CLIP vs DINO in Visual Perception Module }{sec:vpm_exp}
\ToCEntry{0}{Comparison with RL based methods}{sec:rl}
\ToCEntry{0}{Not Just Following Instructions: iSHIFT Understands}{sec:beyond_gt}
\ToCEntry{0}{Limitations}{sec:limitations}
\ToCEntry{0}{Qualitative Results and Comparisons}{sec:qualitative}

\hrule

\section{Dataset Enhancement}
\label{sec:dataset_enhancement}

Algorithm~\ref{alg:training_strategy} describes how each dataset sample is reformatted to include latent thinking and latent perception tokens. For every triplet $(I, T, A)$, we apply a simple rule-based classifier $f_{\text{cat}}$, implemented as an \texttt{if-else} condition, to determine whether the action $A$ requires precise visual grounding. If grounding is required, the sample is routed to the \emph{slow path}; otherwise, it follows the \emph{fast path}.

All the instructions are enriched with a short sequence of latent thinking tokens that supervise implicit reasoning. Samples assigned to the slow path additionally receive a bounded block of latent perception tokens, encouraging the model to rely more heavily on the localized image features. The resulting sequence $\mathcal{S}$, combined with the original image and action label, is added to the enhanced dataset $\mathcal{D}'$.

This process transforms the original dataset into a unified format in which all examples contain lightweight latent-thinking supervision, while only grounding-critical instances include additional perception cues, enabling the model to learn adaptive slow fast reasoning.

\begin{algorithm}[h]
\caption{\textbf{Dataset Enhancement and Training Strategy in iSHIFT}}
\label{alg:training_strategy}
\begin{algorithmic}[1]
\Require Dataset $\mathcal{D} = \{(I, T, A)\}$ where $I$ is the UI screenshot, $T$ is the textual instruction, and $A$ is the ground truth action
\Ensure Enhanced dataset $\mathcal{D}'$ with structured latent-token sequences
\vspace{1pt}

\State Initialize $\mathcal{D}' \gets \emptyset$
\For{each sample $(I, T, A)$ in $\mathcal{D}$}
    \State Determine perception requirement:
    \[
        r = f_{\text{cat}}(A) \in \{\text{Low}, \text{High}\}
    \]

    \If{$r = \text{Low}$}
        \Comment{\textbf{Fast Path (Direct Reasoning)}}
        \State Construct sequence:
        \[
            \mathcal{S} = [T, \texttt{<bot>}, \underbrace{\langle z_1 \rangle, \ldots, \langle z_n \rangle}_{\text{latent thinking tokens}}, \texttt{<eot>}]
        \]
        \State Each $\langle z_i \rangle$ represents an implicit thinking step 
    \Else
        \Comment{\textbf{Slow Path (Perception-Aware Reasoning)}}
        \State Construct sequence:
        \[
        \begin{aligned}
        \mathcal{S} = [&T, \texttt{<bot>}, 
        \underbrace{\langle z_1 \rangle, \ldots, \langle z_n \rangle}_{\text{latent thinking tokens}}, 
        \texttt{<eot>},
        \texttt{<bop>}, \texttt{<ctrl>}, \texttt{<eop>}]
        \end{aligned}
        \]

    \EndIf

    \State Add enhanced sample $(\mathcal{S}, I, A)$ to $\mathcal{D}'$
\EndFor
\vspace{1pt}
\State \Return Enhanced dataset $\mathcal{D}'$
\end{algorithmic}
\end{algorithm}

\begin{figure*}[h!]
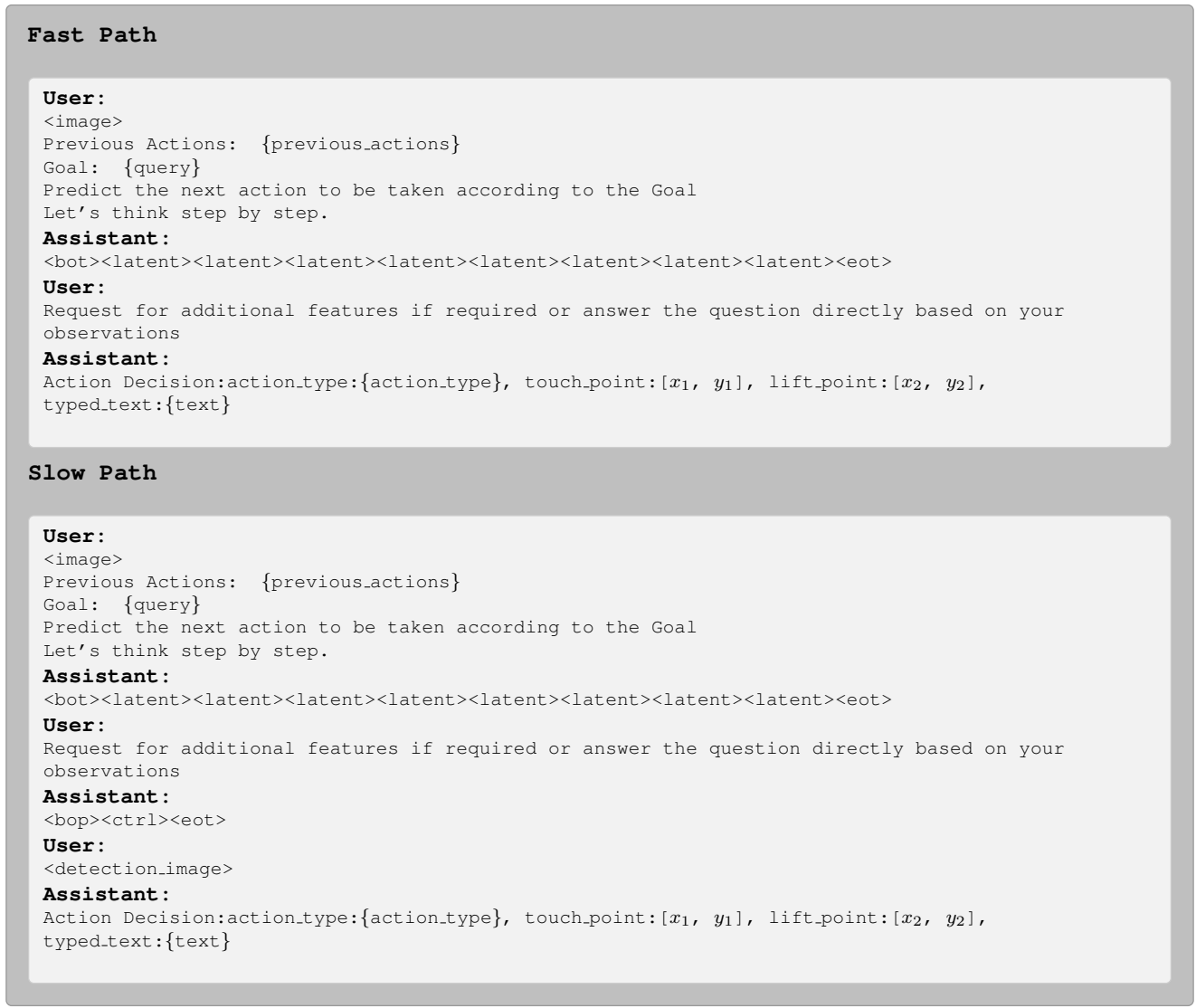

\centering

\begin{tcolorbox}[colback=gray!50, colframe=gray!80,
  boxrule=0.5pt, arc=2pt, left=6pt, right=6pt, top=6pt, bottom=6pt, 
  width=\textwidth]
\small\ttfamily
{\normalsize \textbf{Fast Path}} \\
\vspace{-5pt}
\begin{tcolorbox}[colback=gray!10, colframe=gray!40,
  boxrule=0.5pt, arc=2pt, left=3pt, right=3pt, top=3pt, bottom=3pt,
  width=\textwidth]

{\small\bfseries User:}

{\footnotesize\ttfamily
\noindent\texttt{<image>}\\
Previous Actions: \texttt{\{previous\_actions\}}\\
Goal: \texttt{\{query\}}\\
Predict the next action to be taken according to the Goal\\
Let's think step by step.
\par}

{\small\bfseries Assistant:\par}
{\footnotesize\ttfamily
\noindent\texttt{<bot><latent><latent><latent><latent><latent><latent><latent><latent><eot>}
\par}
{\small\bfseries User:\par}
{\footnotesize\ttfamily
Request for additional features if required or answer the question directly based on your observations
\par}

{\small\bfseries Assistant:\par}
{\footnotesize\ttfamily
Action Decision:action\_type:\{action\_type\}, touch\_point:[$x_1$, $y_1$], lift\_point:[$x_2$, $y_2$], typed\_text:\{text\}\\
\par}

\end{tcolorbox}

{\normalsize \textbf{Slow Path}} \\
\vspace{-5pt}
\begin{tcolorbox}[colback=gray!10, colframe=gray!40,
  boxrule=0.5pt, arc=2pt, left=3pt, right=3pt, top=3pt, bottom=3pt,
  width=\textwidth]

{\small\bfseries User:\par}

{\footnotesize\ttfamily
\noindent\texttt{<image>}\\
Previous Actions: \texttt{\{previous\_actions\}}\\
Goal: \texttt{\{query\}}\\
Predict the next action to be taken according to the Goal\\
Let's think step by step.
\par}

{\small\bfseries Assistant:\par}
{\footnotesize\ttfamily
\noindent\texttt{<bot><latent><latent><latent><latent><latent><latent><latent><latent><eot>}
\par}
{\small\bfseries User:\par}
{\footnotesize\ttfamily
Request for additional features if required or answer the question directly based on your observations
\par}

{\small\bfseries Assistant:\par}
{\footnotesize\ttfamily
\noindent\texttt{<bop><ctrl><eot>}
\par}
{\small\bfseries User:\par}
{\footnotesize\ttfamily
\texttt{<detection\_image>}
\par}

{\small\bfseries Assistant:\par}
{\footnotesize\ttfamily
Action Decision:action\_type:\{action\_type\}, touch\_point:[$x_1$, $y_1$], lift\_point:[$x_2$, $y_2$], typed\_text:\{text\}\\
\par}

\end{tcolorbox}

\end{tcolorbox}

\caption{
\textbf{slow fast Prompt Formatting.}
Illustration of how each dataset sample is reformatted into fast and slow path variants.  
Fast path samples receive only latent thinking tokens, whereas Slow path samples additionally include latent perception tokens to support precise grounding, following the rule-based classifier described in Algorithm~\ref{alg:training_strategy}.
}

\label{fig:prompt}

\end{figure*}

\section{Reproducibility and Implementation Details}
\label{sec:implementation_details}

To facilitate reproducibility and provide clarity on our experimental setup, we detail our training pipeline, hyperparameter choices, prompt template, and datasets used in iSHIFT. Our goal is to ensure that all implementation decisions including model initialization, optimization strategy, action-space normalization, and dataset splits are transparent. \textit{All code, configuration files, and data-processing scripts will be made publicly available upon acceptance.} Below, we outline the computation setup, phase-wise training procedure, and dataset-specific considerations followed in our experiments.

\subsection{Computation and Training Details}
\label{subsec:compute}
Our iSHIFT agent builds on the Qwen2-VL-2B foundation model \cite{qwen2vl}, initialized with its pre-trained weights. The Perception Module uses a frozen DINOv2-L encoder \cite{oquab2024dinov} for stable and efficient feature extraction. To control computational cost in our 2.5B model, the action history is truncated to the two most recent actions. Training is performed on a NVIDIA A100 (80GB) GPU with AdamW and DeepSpeed ZeRO Stage 2. Our training strategy contains three phases:

\begin{enumerate}
    \item \textbf{Alignment Phase:} Only the Visual Perception Module’s cross-attention projector is unfrozen and trained to integrate visual features (learning rate $2\times10^{-3}$, batch size 32, gradient accumulation 4).
    \item \textbf{Thought Training Phase:} The model learns implicit thought
generation using the Android in the Zoo dataset \citep{zhang-etal-2024-android}
(learning rate $3\times10^{-5}$, batch size 16, accumulation~4). Since AITZ provides ground truth thought annotations, we use it
exactly as released without any modification. This stage teaches the latent
tokens their deliberative role, and we observe that the resulting implicit
reasoning generalizes with meaningful thought patterns well beyond the AITZ
domain.
    \item \textbf{Fine-tuning Phase:} Using the thought-trained checkpoint, the model learns the full adaptive slow fast strategy on downstream GUI datasets, with the same batchsize and accumulation settings and a learning rate of $2\times10^{-5}$.
\end{enumerate}

\noindent \textbf{Inference:} During inference, the model receives both the instruction and the screenshot of the current GUI state along with previous actions if any. It then performs implicit task assessment to decide the appropriate reasoning path. If the task is simple, the model follows the fast path and directly predicts the action. If the task requires finer visual grounding, it selects the slow path and outputs the corresponding perception tokens. These tokens trigger the visual perception module, which returns localized features for the queried regions. Using these refined features, the model computes precise grounding coordinates and generates the final action.

\subsection{Hyperparameters}
\label{subsec:hyperparams}
Because iSHIFT is built directly on the Qwen2-VL architecture, nearly all
architectural and optimization hyperparameters are inherited from the base
model, leaving only a single meaningful hyperparameter to tune: the number of
latent thinking tokens. This parameter controls the depth of the model’s
implicit deliberation, and as demonstrated in our ablation on latent token
count (\S \ref{sec:ablations}) has a tangible impact on
performance. Based on this analysis, we set the
number of latent tokens to \textbf{8}, which offers the optimal balance
between reasoning capacity and efficiency.

\subsection{Prompt Template}
\label{subsec:prompt_template}
We employ a unified prompt template that specifies the instruction, screenshot, previous actions, and the latent tokens according to the Dataset Enhancement strategy (\S \ref{sec:dataset_enhancement}) described above. As illustrated in Figure ~\ref{fig:prompt}, the prompt explicitly structures the input into well defined blocks, enabling the model to reason over task goals, past actions, and visual context in a predictable format. This prompt formulation is shared across all training phases and benchmarks, ensuring that iSHIFT can reliably initiate implicit thinking, activate the slow path when needed, and generate grounded actions with minimal ambiguity. 

\subsection{Dataset Details}
\label{subsec:dataset_details}
AITW~\cite{NEURIPS2023_bbbb6308} is a large-scale Android smartphone
interaction environment consisting of approximately 30k natural-language
instructions and 715k trajectories. We obtain our train, validation, and test splits from the AutoUI \cite{zhan2023autoui} benchmark, following the standardized partitioning used in prior work
to ensure consistent and comparable evaluation across models. The action space includes 12 discrete actions: \texttt{CLICK}, \texttt{TYPE}, \texttt{SELECT}, \texttt{SCROLL UP},
\texttt{SCROLL DOWN}, \texttt{SCROLL LEFT}, \texttt{SCROLL RIGHT},
\texttt{PRESS BACK}, \texttt{PRESS HOME}, \texttt{PRESS ENTER},
\texttt{STATUS TASK COMPLETE}, and \texttt{STATUS TASK IMPOSSIBLE}.

Android Control~\cite{li2024effectsdatascaleui} is an Android UI interaction
benchmark that provides 13,604 training episodes (74,722 step-level actions)
and 2,855 test episodes, offering a controlled environment for evaluating
generalization in mobile UI manipulation. We use the official train--test
splits provided by the benchmark to ensure consistency with prior work. The
dataset defines a structured action space covering nine interaction types:
\texttt{click}, \texttt{long\_press}, \texttt{type}, directional
\texttt{scroll}, \texttt{navigate\_home}, \texttt{navigate\_back},
\texttt{open\_app}, \texttt{wait}, and \texttt{terminate}. These actions enable fine-grained control over device navigation, text entry, and task completion signaling. We modify this dataset to match the action space of AITW for consistent training.

GUI Odyssey~\cite{lu2024gui} is a large-scale benchmark designed to evaluate
generalization in multi-device and multi-application GUI interaction. The
dataset contains 8,334 episodes and provides several standardized train--test
configurations. In the Random setup, the data is split
80/20, yielding approximately 6,667 training and 1,667 testing episodes. The
Task subset allocates data across six task categories in a
6:1 ratio, resulting in roughly 7,144 training and 1,190 test episodes. For
the Device configuration, all 1,381 episodes collected on
the Pixel Tablet are reserved for testing, while the remaining 6,953 episodes
form the training set. Finally, the App subset divides episodes
by application frequency with an 85/15 ratio, producing approximately 7,084
training and 1,250 test samples. GUI Odyssey defines a nine-action interaction
space consisting of \texttt{CLICK}, \texttt{LONG\_PRESS}, \texttt{SCROLL},
\texttt{TYPE}, \texttt{COMPLETE}, \texttt{IMPOSSIBLE}, \texttt{HOME} and
\texttt{BACK}, each with structured arguments specifying
positions, gestures, or navigation commands. We modify this dataset to match
the action space of AITW for consistent training and evaluation across
benchmarks.

GUIAct~\cite{chen-etal-2025-guicourse} is a multi-platform benchmark for evaluating  task-oriented GUI agents across web and smartphone environments. The dataset provides these standard splits: {Web-Single, with 67k training and 1.4k testing samples and Smartphone, with 9,157 training and 2k testing samples. The action space is as follows: \texttt{Click}, \texttt{Hover}, \texttt{Tap}, \texttt{Input}, \texttt{Type}, \texttt{Copy}, \texttt{Paste}, \texttt{Scroll}, \texttt{Drag},  \texttt{Swipe}, \texttt{Enter} \texttt{Answer}. For consistent evaluation across benchmarks, we normalize this action space by merging semantically equivalent operations.

\section{Initialization of Latent Tokens}
\label{sec:init}
To validate the importance of our implicit thought–training stage, we conducted an ablation in which the latent thought tokens were initialized from scratch and trained only during downstream fine-tuning, without exposure to the Android in the Zoo thought-supervision phase. As shown in Table~\ref{tab:cot_vs_ours}, removing the thought pretraining leads to consistent performance drops across all evaluation splits, with the largest declines observed in the more open-ended General and G.Apps categories. In contrast, equipping the model with thought pretraining exhibit stronger reasoning consistency and improved robustness, yielding an absolute average gain of 3.3 points. These results confirm that early alignment of latent tokens toward deliberative behavior plays a crucial role in enabling the model to effectively leverage the adaptive slow fast strategy during GUI interaction.

\begin{table}[h]
    \centering
    \setlength{\tabcolsep}{4pt}
    \renewcommand{\arraystretch}{1.5}
    \begin{tabular}{l c c c c c c}
    \hline
    Method & General & Install & G.Apps & WebShop & Single & Avg \\
    \hline
    w/o Thought Pretraining & 66.75 & 78.34 & 67.80 & 68.96 & 83.34 & 73.04 \\
    with Thought Pretraining & \textbf{70.60} & \textbf{80.82} & \textbf{71.64} & \textbf{72.60} & \textbf{86.03} & \textbf{76.34} \\
    \hline
    \end{tabular}
    \caption{\textbf{Effect of latent-token initialization on adaptive slow fast reasoning}. Thought-pretrained latent tokens significantly improve performance, confirming their role in enabling reliable slow-path grounding and overall task success.}
    \label{tab:cot_vs_ours}
\end{table}


\begin{figure*}[t]
    \centering
    \includegraphics[width=\linewidth]{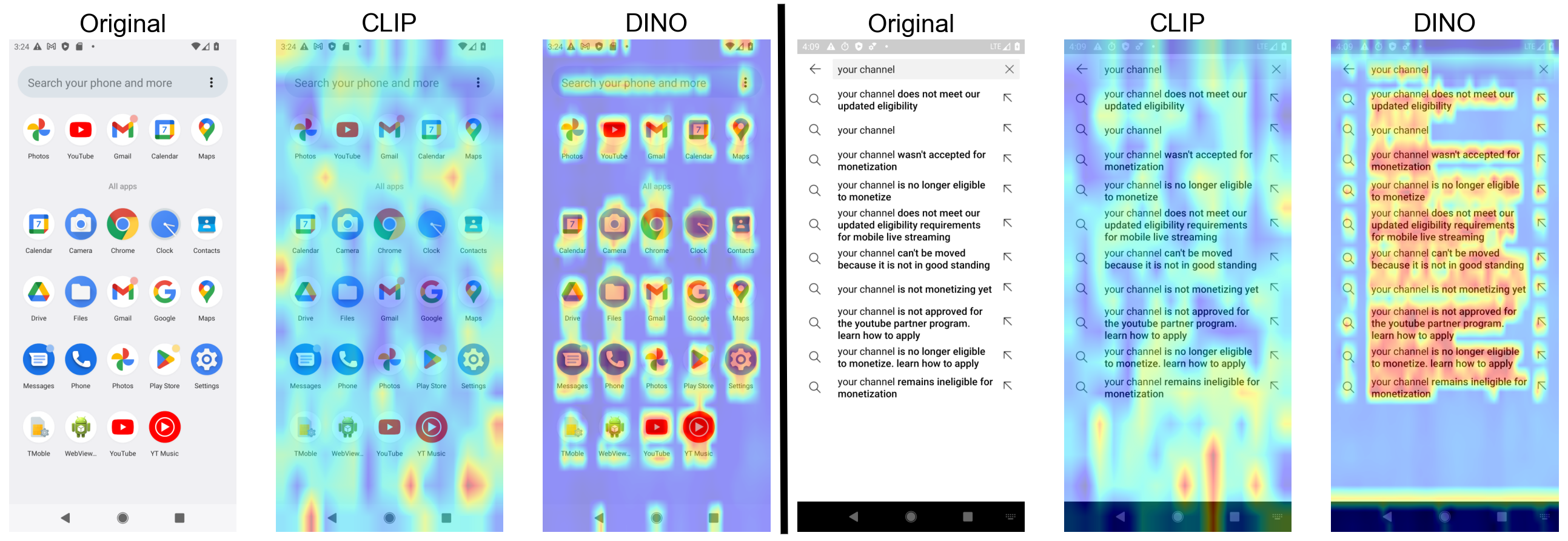}
    \caption{Attention map comparison between CLIP and DINO visual encoders. DINO produces more localized and structured attention over relevant UI elements, while CLIP exhibits broader, less focused activations.}
    \label{fig:encoder_maps}
\end{figure*}

\section{Chain of Thought v/s Latent Thinking}
\label{sec:cot}
To evaluate the effectiveness and generalization ability of our implicit-thinking approach relative to explicit chain-of-thought (CoT) reasoning, we train on the Android In The Zoo (AITZ) dataset \cite{zhang-etal-2024-android} and assess performance on the diverse test splits of Android In The Wild (AITW) \cite{NEURIPS2023_bbbb6308}. AITZ is constructed from a small subset of AITW and is a publicly available dataset that provides ground truth thought annotations. Therefore, for our CoT baseline, we use the dataset exactly as released without modifying or augmenting any annotation. AITZ contains approximately 18k training examples with detailed thoughts and screen descriptions, while the combined AITW test sets include around 127k samples, forming a challenging out-of-distribution benchmark for assessing generalization of the latent tokens. As shown in Table~\ref{tab:cotvslatent}, implicit thinking consistently outperforms explicit CoT across all AITW splits, achieving gains of 8–11 points on average despite using nearly $8\times$ fewer tokens. Notably, the implicit model delivers the strongest improvements in the most complex domains: General, Google Apps, and WebShop, demonstrating that compact latent reasoning not only generalizes better but also provides significantly lower inference overhead compared to explicit CoT.

\begin{table}[h]
    \centering
    \small
    \setlength{\tabcolsep}{3pt}
    \renewcommand{\arraystretch}{1.2}
    \begin{tabular}{l c c c c c c}
    \hline
    Method & General & Install & G.Apps & WebShop & Single & Avg Tokens \\
    \hline
    Chain of Thought & 46.03 & 64.97 & 49.81 & 49.78 & 66.57 & 62 \\
    Implicit Thinking & \textbf{56.54} & \textbf{71.73} & \textbf{60.28} & \textbf{58.75} & \textbf{77.22} & 8 \\
    \hline
    \end{tabular}
    \caption{\textbf{Implicit thinking vs. chain-of-thought when trained on AITZ and evaluated on AITW}. Using AITZ (18k thought annotated samples) for training and the full AITW benchmark (~127k test samples) for evaluation, implicit reasoning delivers consistent improvements over explicit CoT across all domains, while operating with an order-of-magnitude fewer tokens.}
    \label{tab:cotvslatent}
    \vspace{-10pt}
\end{table}

\section{CLIP vs DINO in Visual Perception Module}
\label{sec:vpm_exp}
Figure \ref{fig:encoder_maps} compares attention maps of CLIP and DINO on examples from AITW dataset. As observed, CLIP’s heatmaps show more diffuse, global attention that loosely corresponds to semantically meaningful areas (like icons or text regions), reflecting its image text training objective. In contrast, DINO produces sharper, object centered attention, precisely outlining icons, buttons, and text lines, demonstrating its strong spatial and structural awareness learned through self supervised vision training. Overall, CLIP captures semantic meaning, while DINO captures visual structure and object boundaries.

\section{Comparison with RL based Methods}
\label{sec:rl}
We additionally compare iSHIFT against recent reinforcement learning based GUI agents across multiple benchmarks to contextualize its performance relative to policy optimization approaches. As shown in Table~\ref{tab:aitw_rl_results}, iSHIFT matches or surpasses strong RL baselines such as DigiRL and DistRL on the custom AITW test subsets, despite using a purely supervised training pipeline without any environment rollouts or reward engineering. Notably, our model provides large gains on the Web Shopping split, highlighting the benefits of adaptive perception in visually dense, multi step tasks. Beyond AITW, iSHIFT also achieves substantial improvements on the Android Control and GUI Odyssey benchmarks (Table~\ref{tab:rl_results_2}), outperforming prior RL agents including UI-R1, GUI-R1, and SWIRL, by wide margins across both high and low complexity scenarios. These results emphasize that our token based slow fast strategy yields stronger generalization and more reliable action grounding than methods relying on reinforcement learning.

\begin{table*}[h]
\centering
\small
\begin{minipage}[t]{0.48\linewidth}
\centering
\setlength{\tabcolsep}{4pt}
\renewcommand{\arraystretch}{1}
\begin{tabular}{l c c c}
\hline
\textbf{Method} & General & Web Shopping & Test Set \\
\hline
DigiRL & 71.90 & 67.20 & \multirow{2}{*}{First 96 task samples} \\
Ours   & \textbf{72.25} & \textbf{72.28} & \\
\hline
DistRL & \textbf{73.20} & 68.50 & \multirow{2}{*}{First 128 task samples} \\
Ours   & 71.85 & \textbf{73.19} & \\
\hline
\end{tabular}
\caption{\textbf{Comparison with RL based agents on AITW}. Results on the custom AITW test subsets used by DigiRL and DistRL. iSHIFT consistently matches or surpasses RL baselines despite using purely supervised training without rollouts or rewards.}
\label{tab:aitw_rl_results}
\end{minipage}
\hfill
\begin{minipage}[t]{0.48\linewidth}
\centering
\footnotesize
\setlength{\tabcolsep}{4pt}
\renewcommand{\arraystretch}{1.25}
\begin{tabular}{l c c c}
\hline
\multirow{2}{*}{\textbf{Method}} &
\multicolumn{2}{c}{\textbf{Android Control}} &
\multirow{2}{*}{\parbox{1.1cm}{\centering \textbf{GUI Odyssey}}} \\
& High & Low & \\
\hline
UI-R1      & 45.44 & 66.44 & 32.49 \\
GUI-R1 3B  & 46.55 & 64.41 & 41.33 \\
GUI-R1 7B  & 51.67 & 66.52 & 38.79 \\
SWIRL      & 51.24 & 78.81 & 51.65 \\
\hline
\textbf{iSHIFT} & \textbf{65.6} & \textbf{87.7} & \textbf{73.97} \\
\hline
\end{tabular}
\caption{\textbf{Comparison of iSHIFT with prior RL-based GUI agents on Android Control and GUI Odyssey}. iSHIFT achieves the highest performance across all settings, outperforming UI-R1, GUI-R1 (3B and 7B), and SWIRL in both simple and complex interaction scenarios.}
\label{tab:rl_results_2}
\end{minipage}
\end{table*}

\begin{figure*}[t]
    \centering
    \includegraphics[width=\linewidth]{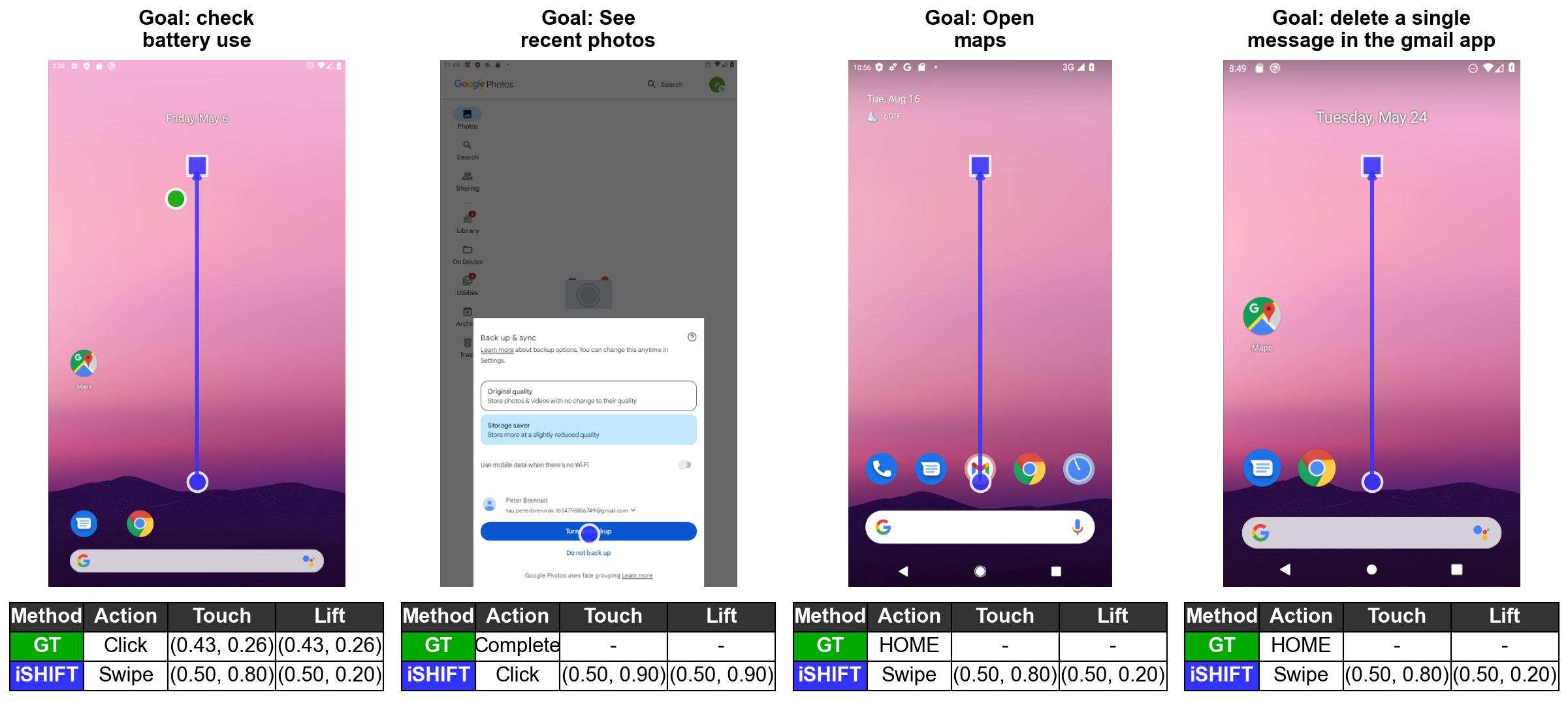}
    \caption{Examples where the ground truth annotations are incorrect, while iSHIFT performs the correct action to achieve the goal.}
    \label{fig:gt_wrong}
\end{figure*}

\section{Not Just Following Instructions: iSHIFT Understands}
\label{sec:beyond_gt}
This section presents qualitative examples from AITW dataset where iSHIFT’s actions differ from annotated ground truth trajectories yet still achieve the intended goal effectively. These examples demonstrate that iSHIFT is not limited to imitating annotations but can flexibly adapt its interactions based on visual context and task objectives. We categorize such behaviors into three representative types: (i) Cases where the provided annotation is wrong while iSHIFT performs the correct action. (ii) iSHIFT takes shorter and more efficient completion routes, and (iii) iSHIFT takes alternate but valid interaction paths that reach the goal. Together, these examples highlight iSHIFT’s robustness and ability to generalize beyond the exact sequences observed during supervision.

\noindent \textbf{Handling Annotation Errors}
Figure \ref{fig:gt_wrong} highlights situations where iSHIFT performs the correct interaction even when the annotated ground truth is incomplete or inaccurate. In tasks such as “Check battery use,” “Open maps,” and “Delete a single message in Gmail,” the annotations specify incorrect or missing actions (e.g., HOME or Click), whereas iSHIFT executes valid gestures that reveal the intended interface or open the correct app. Likewise, in “See recent photos,” iSHIFT selects the appropriate navigation button, while the annotation prematurely marks the task as complete. These examples suggest that minor inconsistencies in GUI annotations do not prevent iSHIFT from executing valid, goal-directed interactions.

\noindent \textbf{More Efficient Task Completions}
As shown in Figure \ref{fig:faster_paths}, iSHIFT often completes tasks through shorter and more efficient action sequences. For instance, in “Turn off JavaScript in Chrome,” it directly clicks the Chrome icon instead of performing an intermediate swipe. In “Check the news” examples, it bypasses the step of opening the app drawer and interacts directly with the Google search bar, selecting an existing suggestion instead of typing a full query. Such behaviors demonstrate that iSHIFT can streamline interactions to minimize unnecessary steps while still reaching the correct goal.

\begin{figure*}[h]
    \centering
    \includegraphics[width=\linewidth]{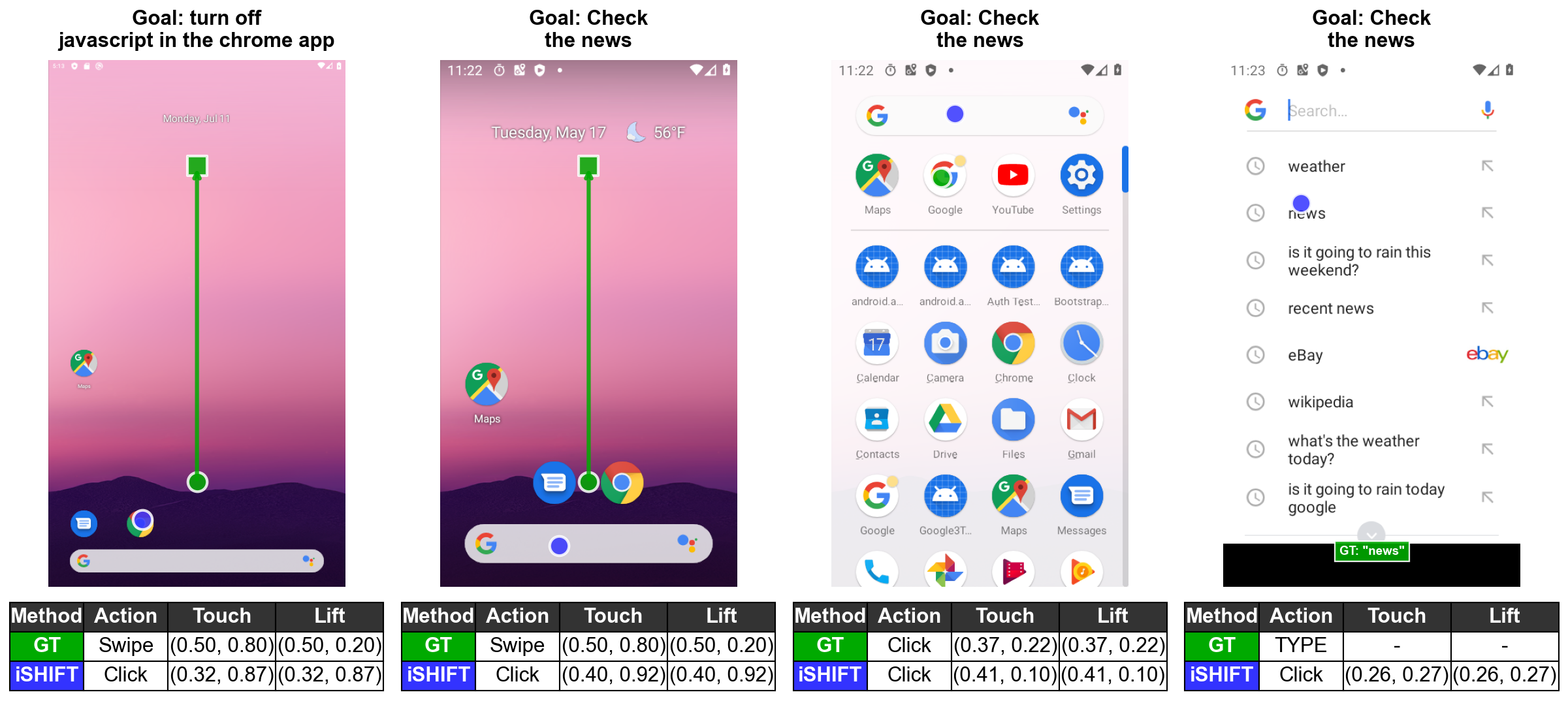}
    \caption{Examples where iSHIFT completes tasks through shorter or more efficient interaction paths than the ground truth trajectories, demonstrating its ability to identify faster yet valid routes to achieve the same goals.}
    \label{fig:faster_paths}
\end{figure*}

\noindent \textbf{Alternate but Valid Paths}
Figure \ref{fig:alternate_path} shows cases where iSHIFT follows a different yet valid sequence of actions to complete the task. For “What’s the news in Chile?”, it clicks on a suggested query instead of pressing ENTER, producing the same result. For “Open a new tab in Chrome,” it accesses the tabs option from the menu rather than using the menu. Similarly, in the Costco and eBay tasks, it selects the search symbol instead of selecting the nearby search suggestions that lead to equivalent outcomes. These examples indicate that iSHIFT can flexibly adapt its interaction pattern while remaining consistent with the task objective.

\begin{figure*}[t]
    \centering
    \includegraphics[width=\linewidth]{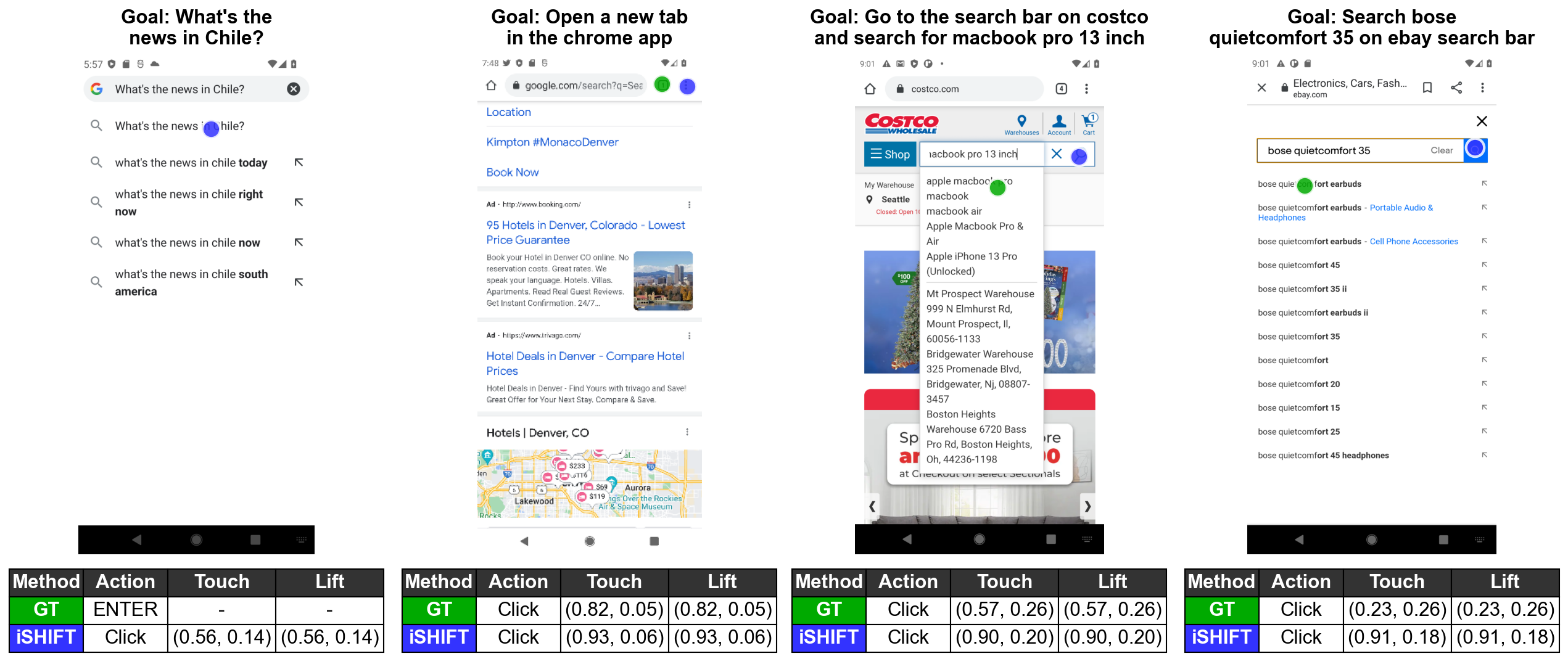}
    \caption{Examples where iSHIFT follows alternate but valid interaction paths that differ from the ground truth actions yet successfully achieve the intended goals.}
    \label{fig:alternate_path}
\end{figure*}

\begin{figure*}
    \centering
    \includegraphics[width=\linewidth]{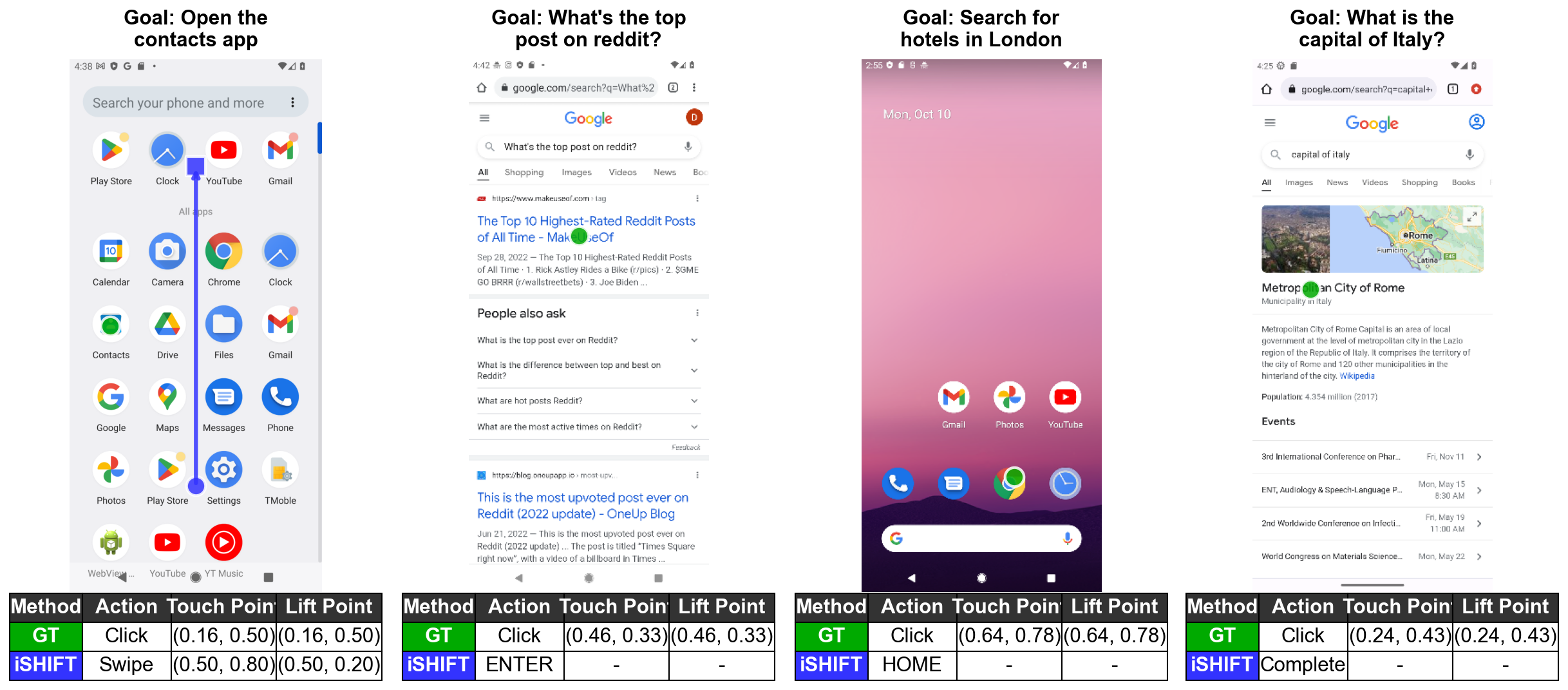}
    \caption{Examples where iSHIFT occasionally selects the fast path for tasks that require deeper perception.}
    \label{fig:choose-fast-instead-of-slow}
\end{figure*}

\section{Limitations}
\label{sec:limitations}
In some cases, iSHIFT performs actions that differ from both the ground truth and the intended goal. These deviations typically stem from: (i) the model occasionally selects the fast path when a task benefits from deeper visual reasoning (see Figure \ref{fig:choose-fast-instead-of-slow}), (ii) it may opt for the slow path on simpler interactions (see Figure \ref{fig:choose-slow-instead-of-fast}), and (ii) the Visual Perception Module may occasionally provide slightly imprecise localized features (see Figure \ref{fig:vpm-error}). Nonetheless, as shown in Figure \ref{fig:aitw_action_dist} and \ref{fig:others_action_dist}, such occurrences are rare, indicating that iSHIFT’s adaptive control and perception mechanisms remain stable.

\begin{figure*}[t]
    \centering
    \includegraphics[width=\linewidth]{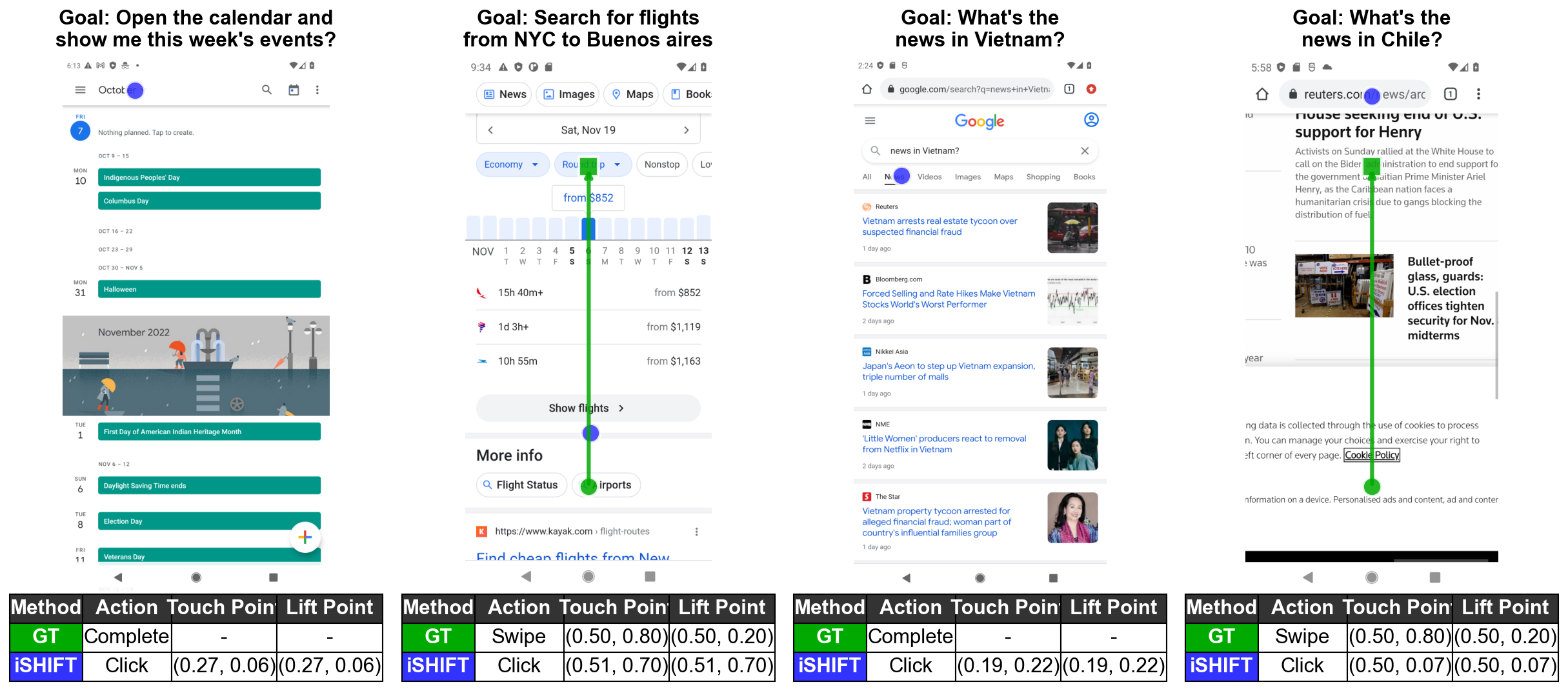}
    \caption{Examples where iSHIFT occasionally selects the slow path for simpler interactions.}
    \label{fig:choose-slow-instead-of-fast}
\end{figure*}

\begin{figure*}[h]
    \centering
    \includegraphics[width=\linewidth]{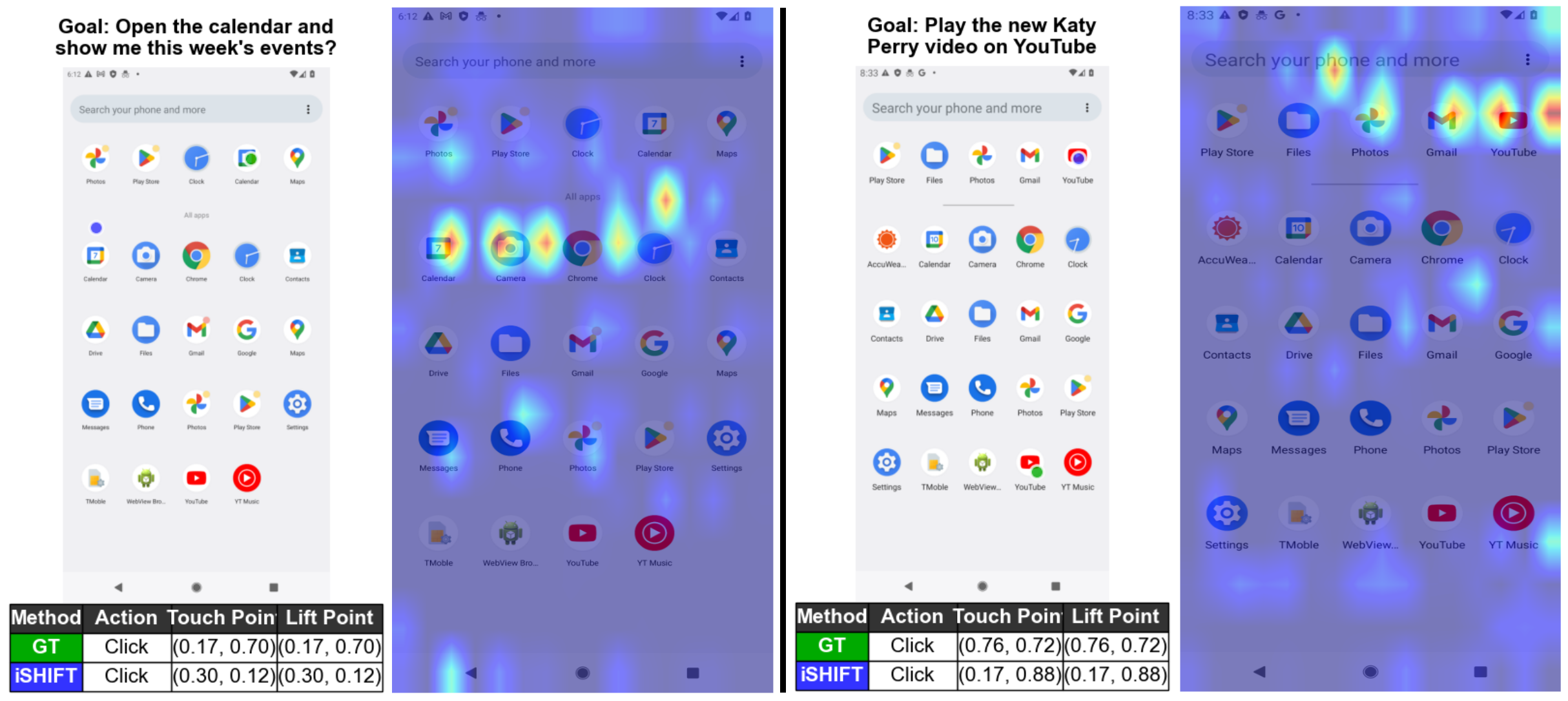}
    \caption{Examples showing minor localization variations from iSHIFT’s Visual Perception Module.}
    \label{fig:vpm-error}
\end{figure*}

\noindent \textbf{Choosing the Fast Path instead of Slow and vice-versa.} 
While iSHIFT is designed to dynamically balance between the slow and fast reasoning paths, occasional mismatches occur where it selects the fast route for tasks requiring deeper perception (see Figure \ref{fig:choose-fast-instead-of-slow}) or the slow route for simpler interactions (see Figure \ref{fig:choose-slow-instead-of-fast}). These cases arise from subtle variations in visual complexity estimation or perception cues. However, as shown in Figure \ref{fig:aitw_action_dist} and \ref{fig:others_action_dist}, such instances are infrequent, iSHIFT maintains a nearly even and well-calibrated distribution of slow and fast path usage, closely aligning with the ground truth statistics. This indicates that the adaptive control mechanism is generally stable, with only minor room for refinement in boundary cases.

\noindent \textbf{Visual Perception Module. }
Figure \ref{fig:vpm-error} illustrates cases where iSHIFT’s Visual Perception Module shows minor localization differences. In the first example, when opening the Calendar app, iSHIFT’s attention is slightly offset from the target icon, leading to a small spatial error. In the second example, when asked to play a YouTube video, it selects one of the two YouTube icons present on the screen, different from the annotated ground truth but still a valid choice.

\begin{figure*}[t]
    \centering
    \includegraphics[width=0.85\textwidth]{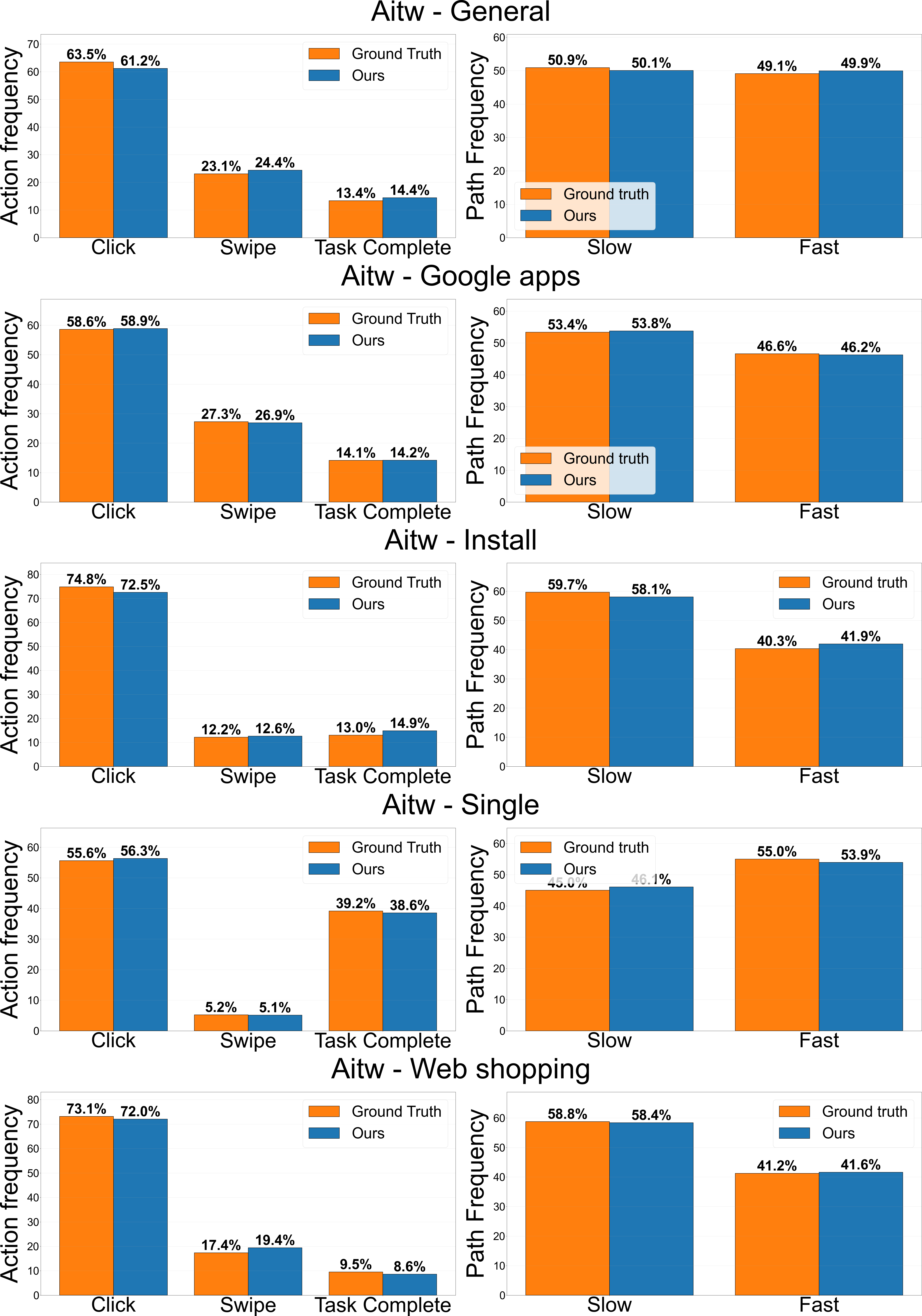}
    \caption{Action and path distribution on AITW test data. iSHIFT closely matches ground truth action frequencies and accurately predicts both slow and fast paths across all task categories.}
    \label{fig:aitw_action_dist}
\end{figure*}

\begin{figure*}
    \centering
    \includegraphics[width=0.85\textwidth]{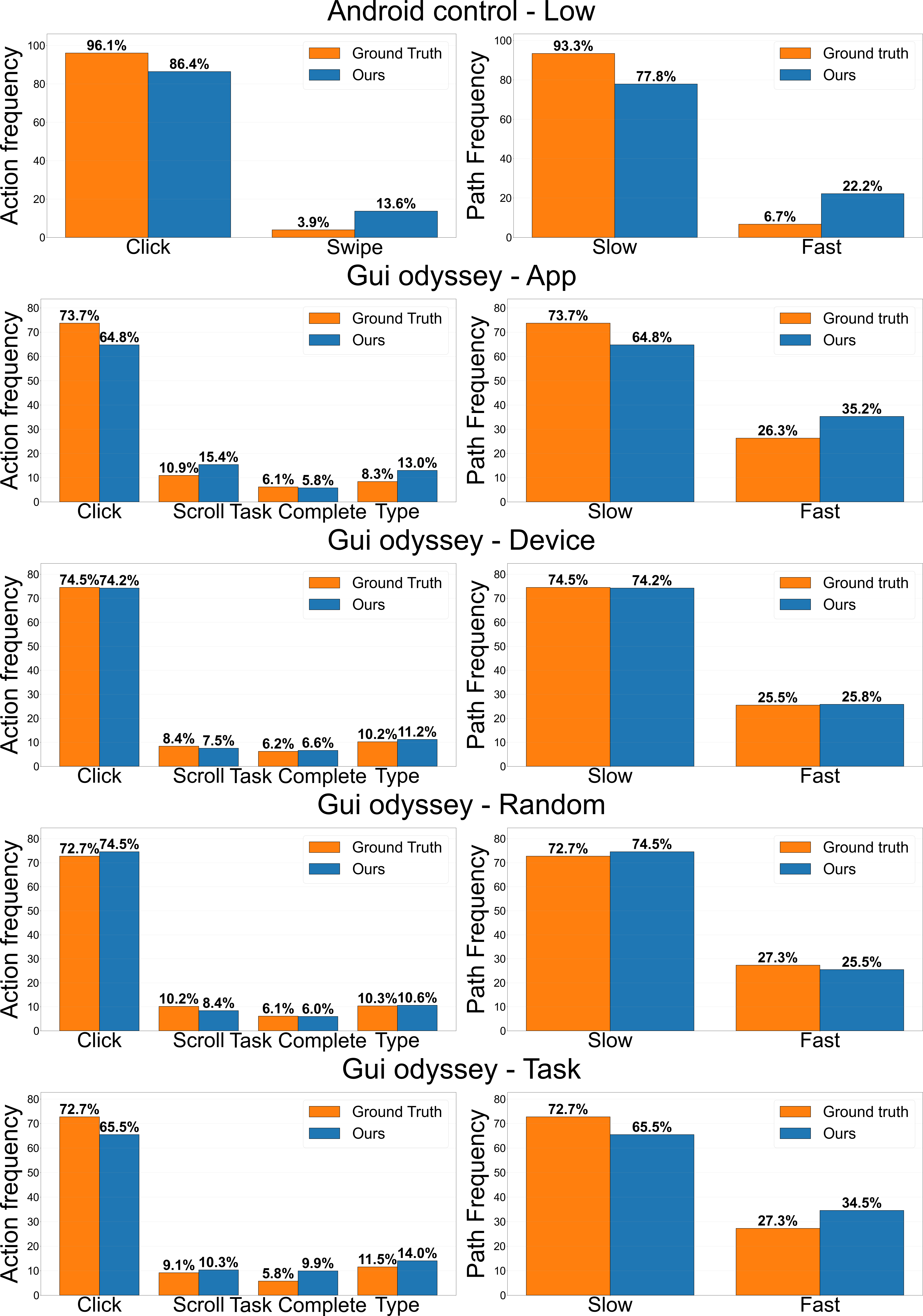}
    \caption{Action and path distribution on Android Control and GUI odyssey test data. iSHIFT closely matches ground truth action frequencies and accurately predicts both slow and fast paths across all task categories.}
    \label{fig:others_action_dist}
\end{figure*}

\section{Qualitative Results and Comparisons}
\label{sec:qualitative}
We provide qualitative examples illustrating how iSHIFT behaves across diverse GUI environments and interaction challenges (Figures \ref{fig:episodes_0} - \ref{fig:episodes_8}). As shown in our qualitative figures, the model reliably selects between the fast and slow paths and produces actions with high spatial precision. We also compare iSHIFT against prior agents such as ShowUI\cite{showui} and SeeClick\cite{cheng2024seeclick} highlighting fewer failure modes and more stable behavior in cases involving dense UI elements, ambiguous affordances, or fine-grained interaction requirements (Figures \ref{fig:comparisons_0} - \ref{fig:comparisons_3}). These examples offer an intuitive understanding of why our adaptive slow fast strategy yields strong improvements across benchmarks.

\begin{figure*}
    \centering
    \includegraphics[width=0.80\textwidth]{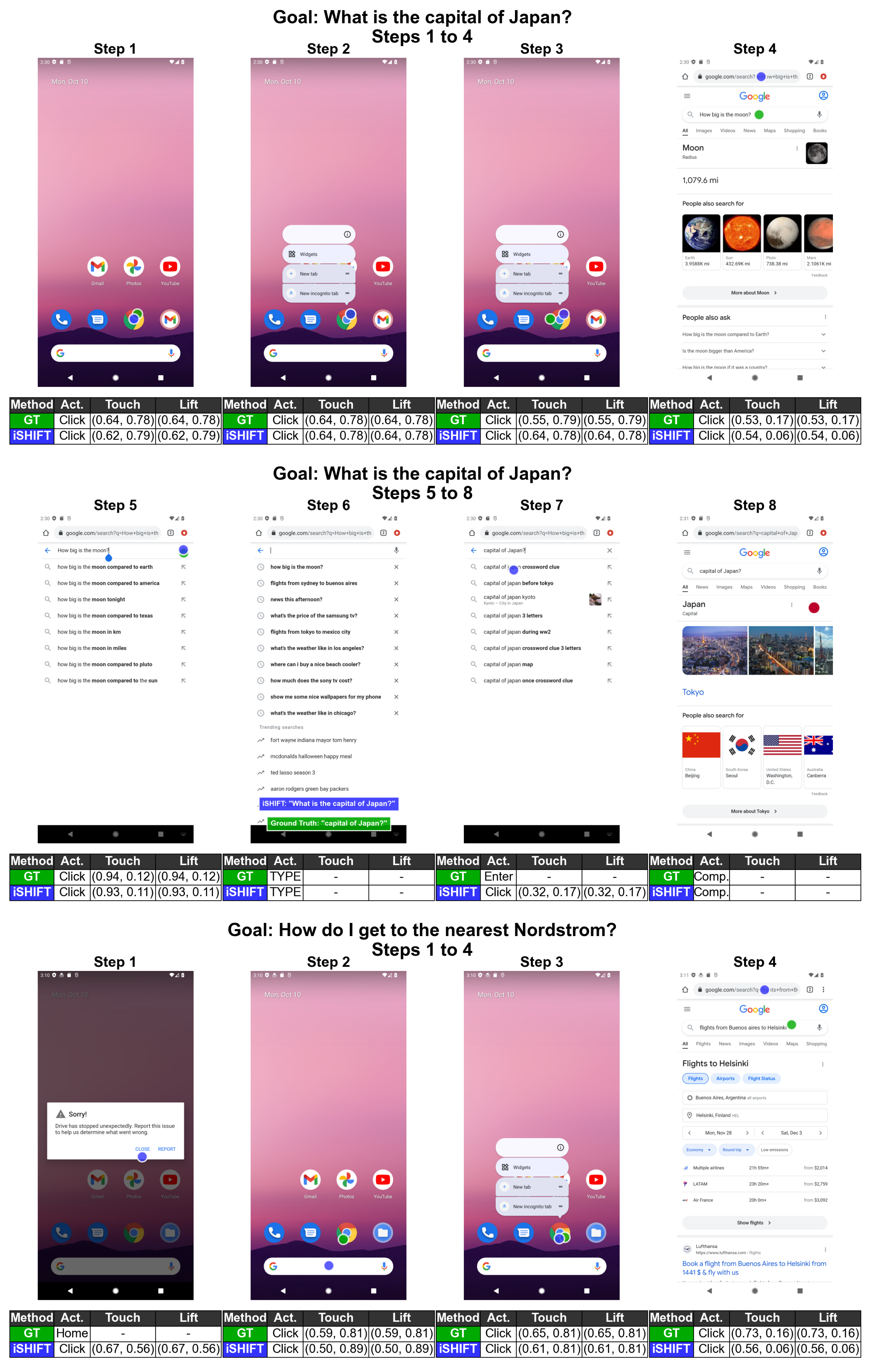}
    \vspace{-10pt}
    \caption{Episodes from the AITW dataset comparing \textcolor{blue}{iSHIFT} trajectories with the \textcolor{Green}{ground-truth} demonstrations. iSHIFT closely follows the intended interaction flow, and when deviations occur, the model selects alternative but valid action sequences that still achieve the correct task outcome.}
    \label{fig:episodes_0}
\end{figure*}
\begin{figure*}
    \centering
    \includegraphics[width=0.80\textwidth]{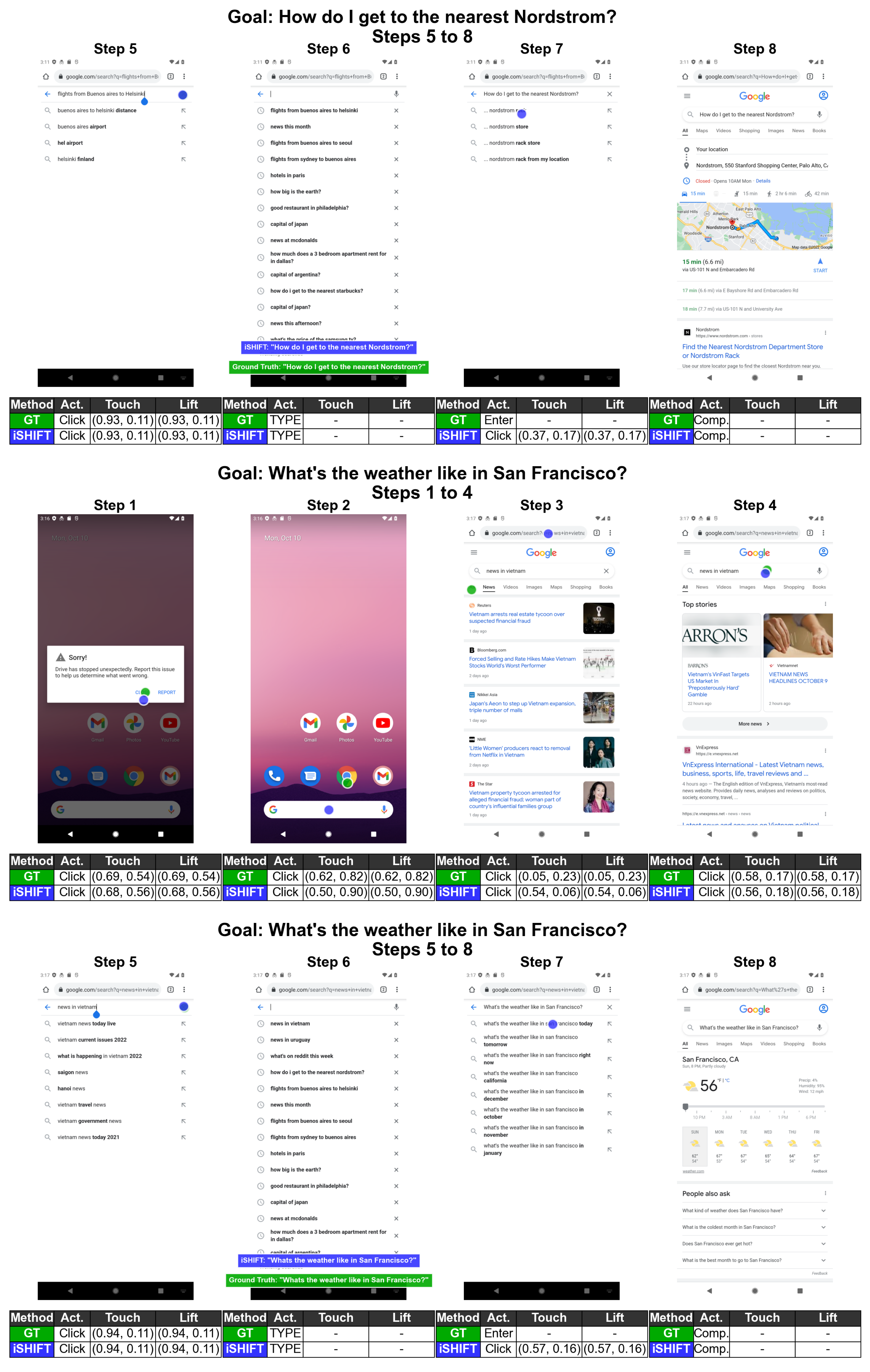}
    \vspace{-10pt}
    \caption{Episodes from the AITW dataset comparing \textcolor{blue}{iSHIFT} trajectories with the \textcolor{Green}{ground-truth} demonstrations. iSHIFT closely follows the intended interaction flow, and when deviations occur, the model selects alternative but valid action sequences that still achieve the correct task outcome.}
    \label{fig:episodes_1}
\end{figure*}

\begin{figure*}
    \centering
    \includegraphics[width=0.80\textwidth]{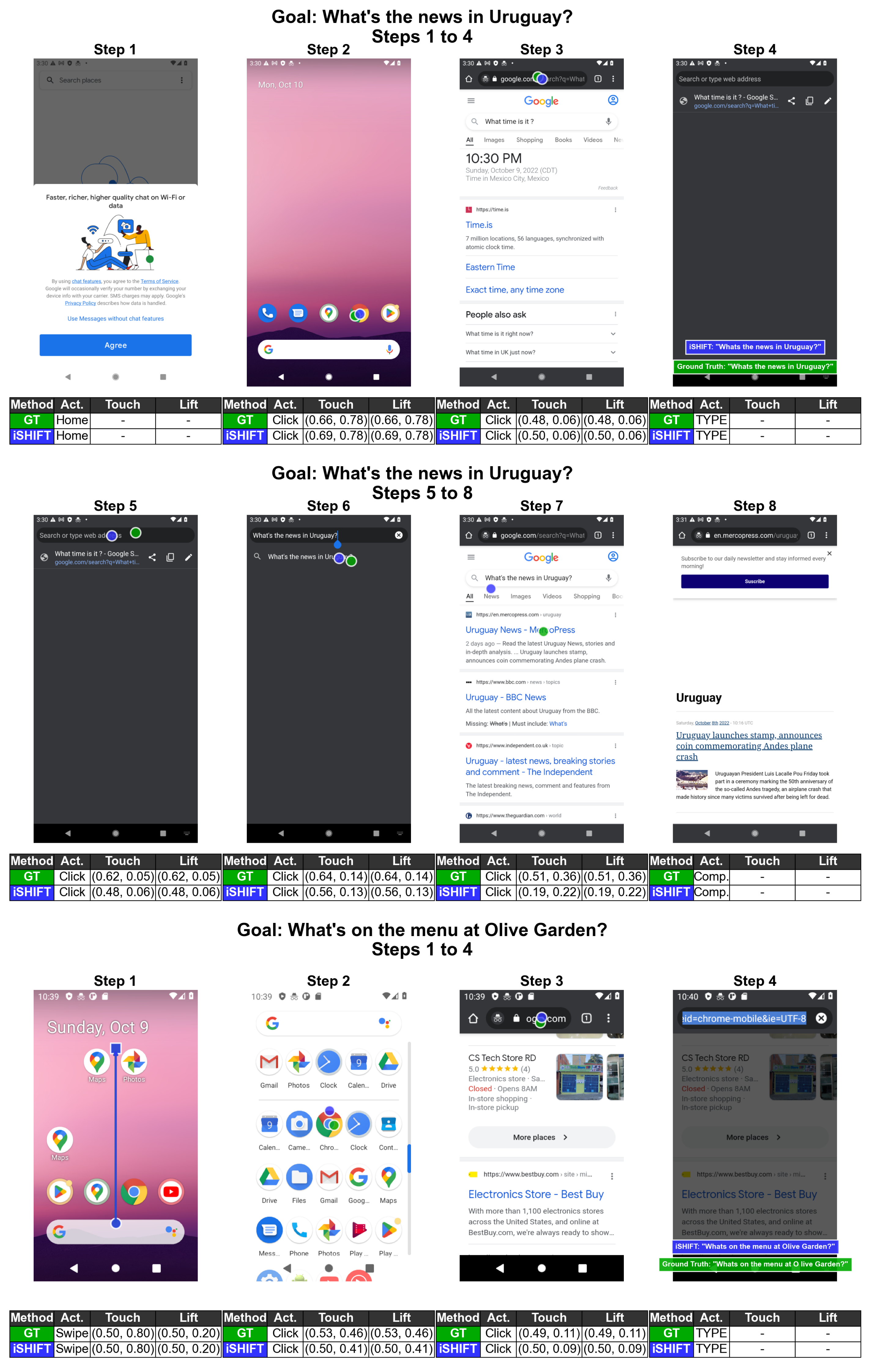}
    \vspace{-10pt}
    \caption{Episodes from the AITW dataset comparing \textcolor{blue}{iSHIFT} trajectories with the \textcolor{Green}{ground-truth} demonstrations. iSHIFT closely follows the intended interaction flow, and when deviations occur, the model selects alternative but valid action sequences that still achieve the correct task outcome.}
    \label{fig:episodes_2}
\end{figure*}

\begin{figure*}
    \centering
    \includegraphics[width=0.80\textwidth]{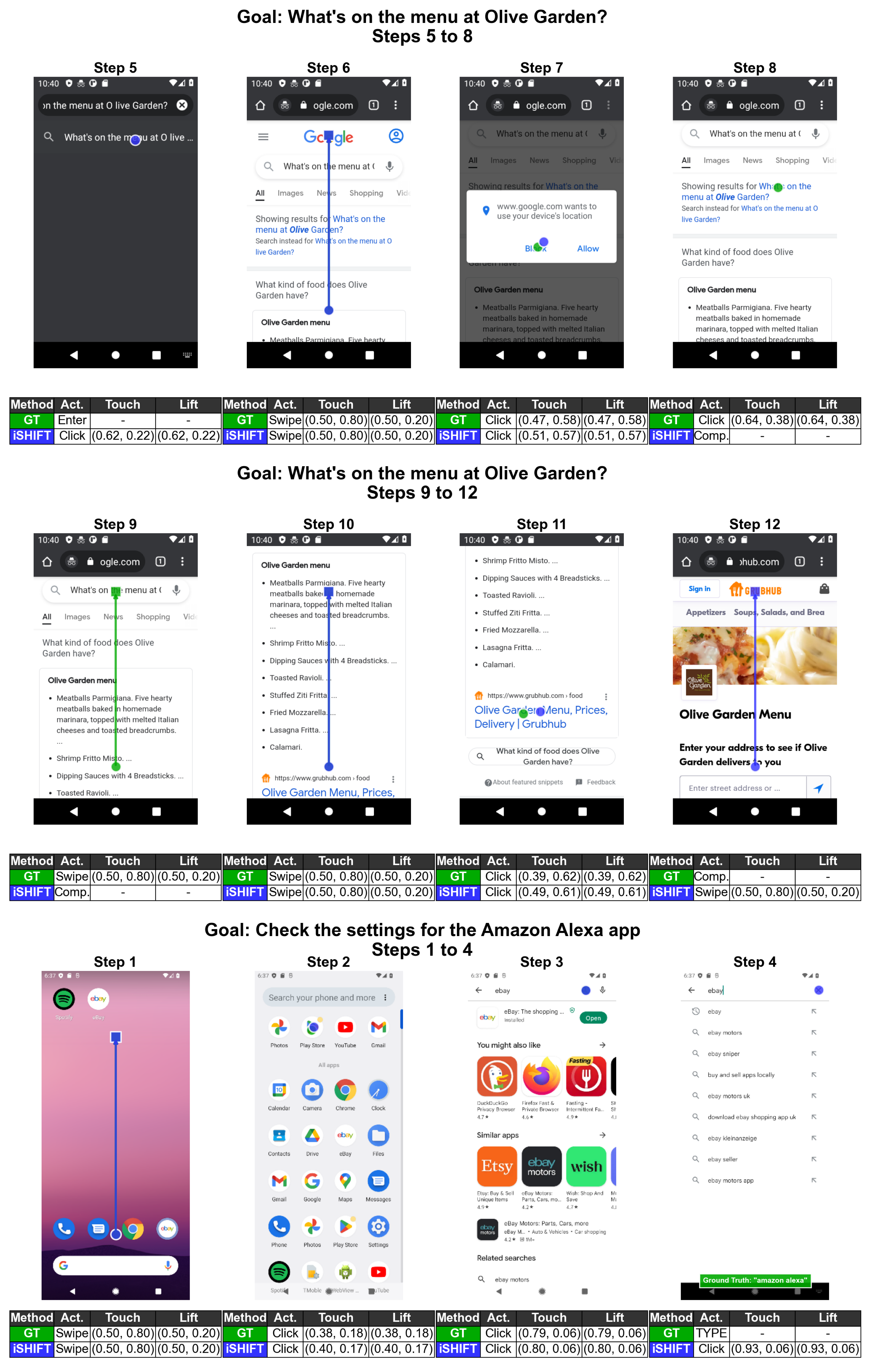}
    \vspace{-10pt}
    \caption{Episodes from the AITW dataset comparing \textcolor{blue}{iSHIFT} trajectories with the \textcolor{Green}{ground-truth} demonstrations. iSHIFT closely follows the intended interaction flow, and when deviations occur, the model selects alternative but valid action sequences that still achieve the correct task outcome.}
    \label{fig:episodes_3}
\end{figure*}

\begin{figure*}
    \centering
    \includegraphics[width=0.80\textwidth]{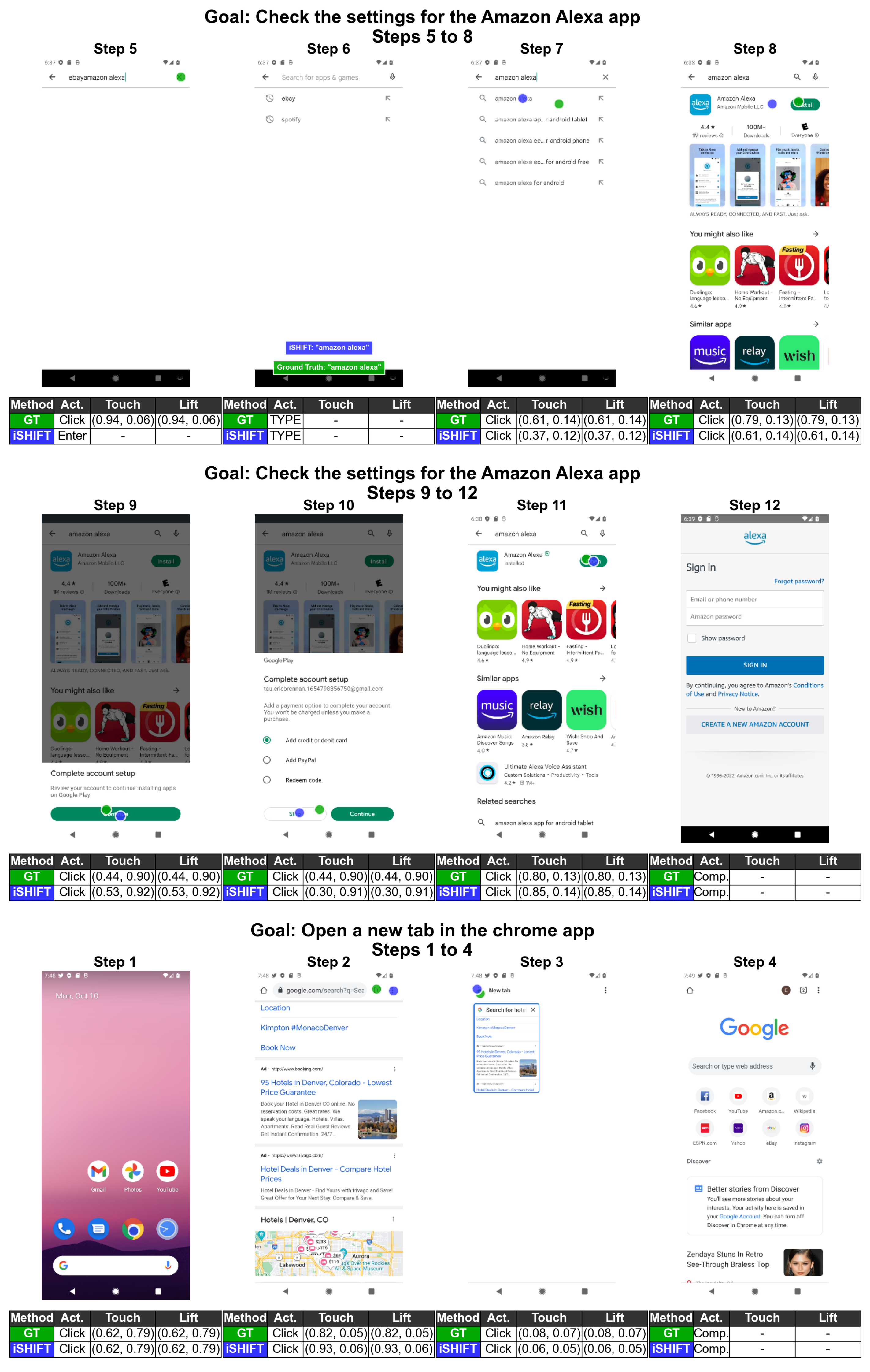}
    \vspace{-10pt}
    \caption{Episodes from the AITW dataset comparing \textcolor{blue}{iSHIFT} trajectories with the \textcolor{Green}{ground-truth} demonstrations. iSHIFT closely follows the intended interaction flow, and when deviations occur, the model selects alternative but valid action sequences that still achieve the correct task outcome.}
    \label{fig:episodes_4}
\end{figure*}

\begin{figure*}
    \centering
    \includegraphics[width=0.80\textwidth]{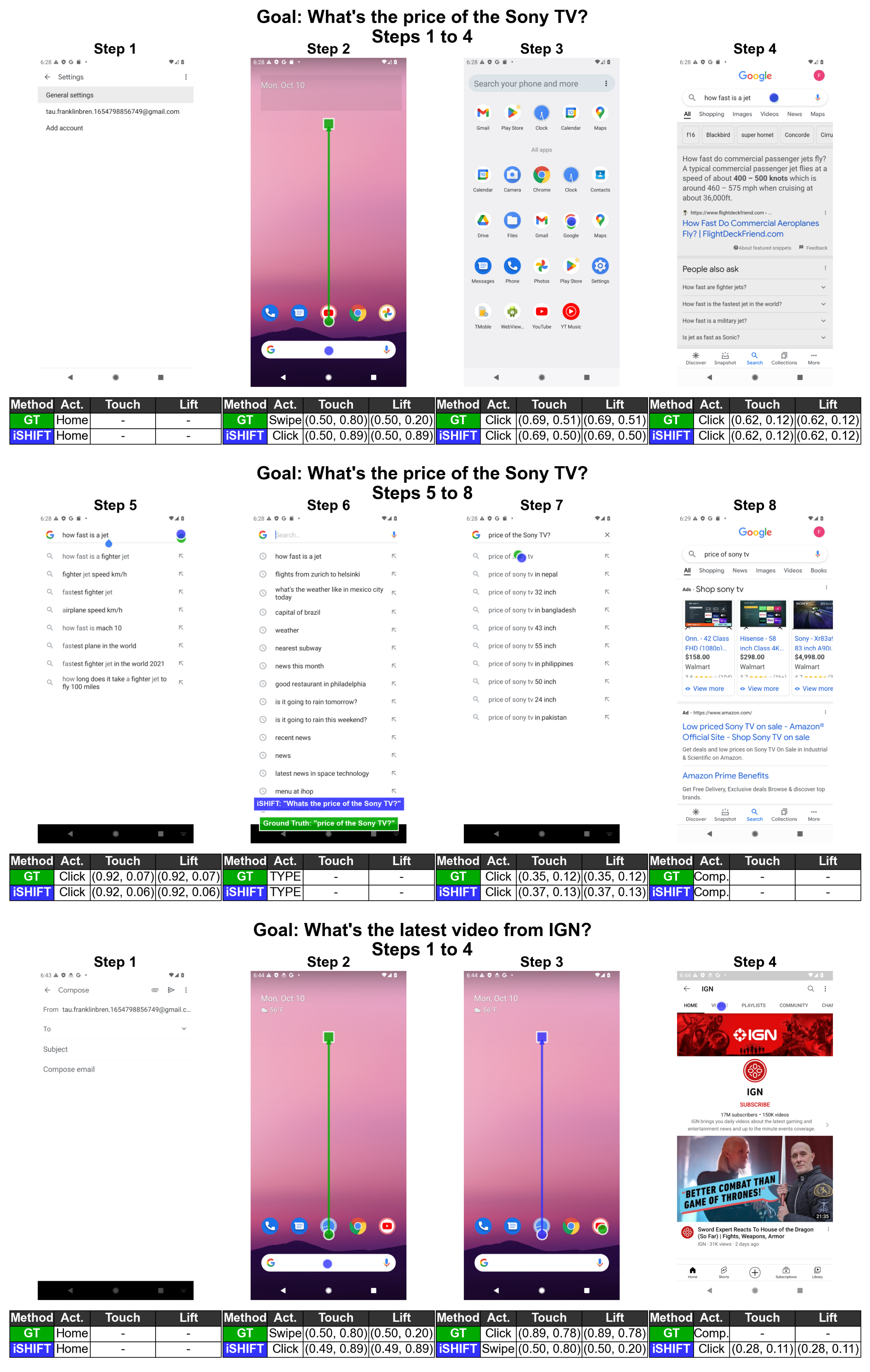}
    \vspace{-10pt}
    \caption{Episodes from the AITW dataset comparing \textcolor{blue}{iSHIFT} trajectories with the \textcolor{Green}{ground-truth} demonstrations. iSHIFT closely follows the intended interaction flow, and when deviations occur, the model selects alternative but valid action sequences that still achieve the correct task outcome.}
    \label{fig:episodes_5}
\end{figure*}

\begin{figure*}
    \centering
    \includegraphics[width=0.80\textwidth]{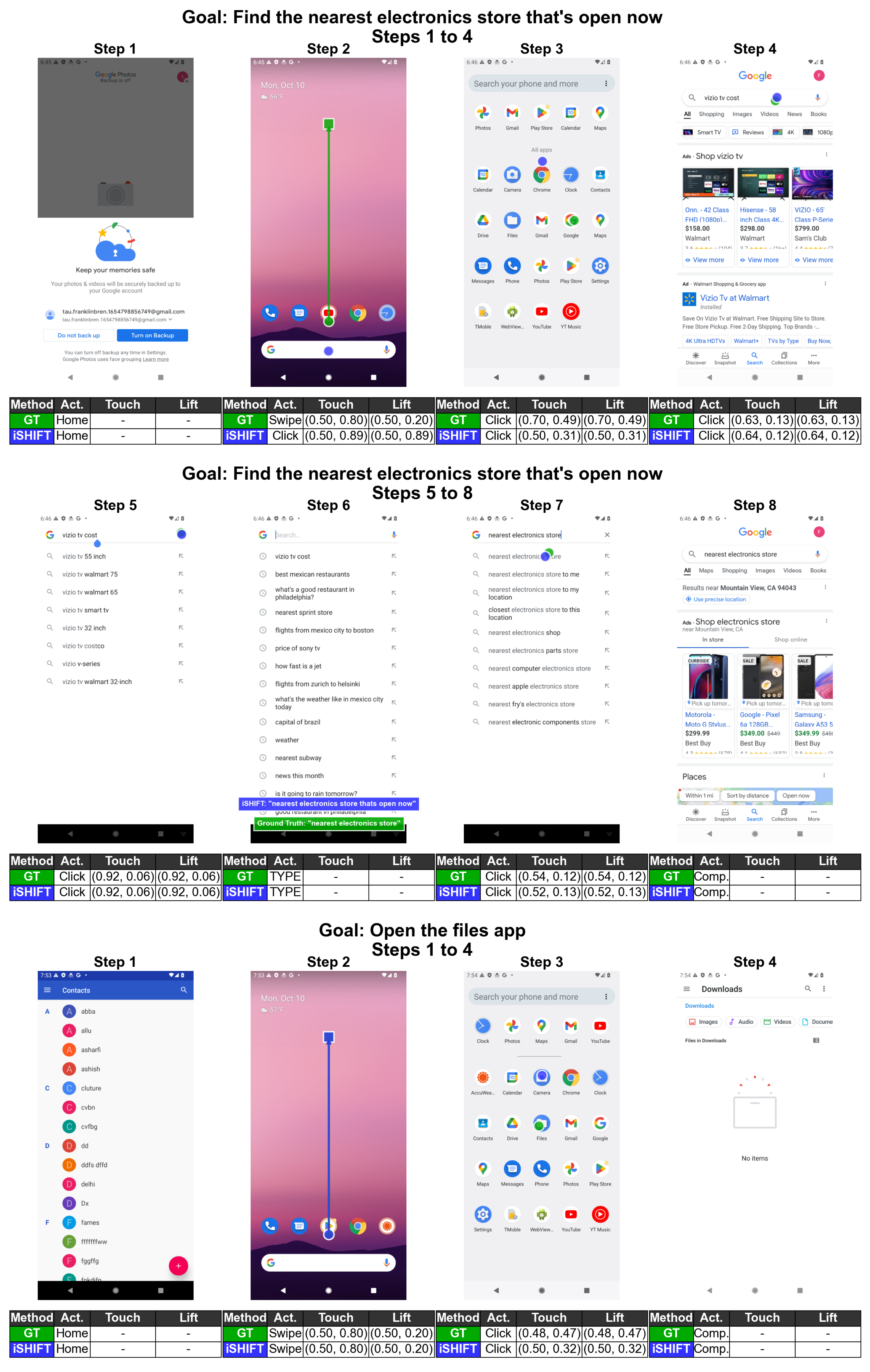}
    \vspace{-10pt}
    \caption{Episodes from the AITW dataset comparing \textcolor{blue}{iSHIFT} trajectories with the \textcolor{Green}{ground-truth} demonstrations. iSHIFT closely follows the intended interaction flow, and when deviations occur, the model selects alternative but valid action sequences that still achieve the correct task outcome.}
    \label{fig:episodes_6}
\end{figure*}

\begin{figure*}
    \centering
    \includegraphics[width=0.80\textwidth]{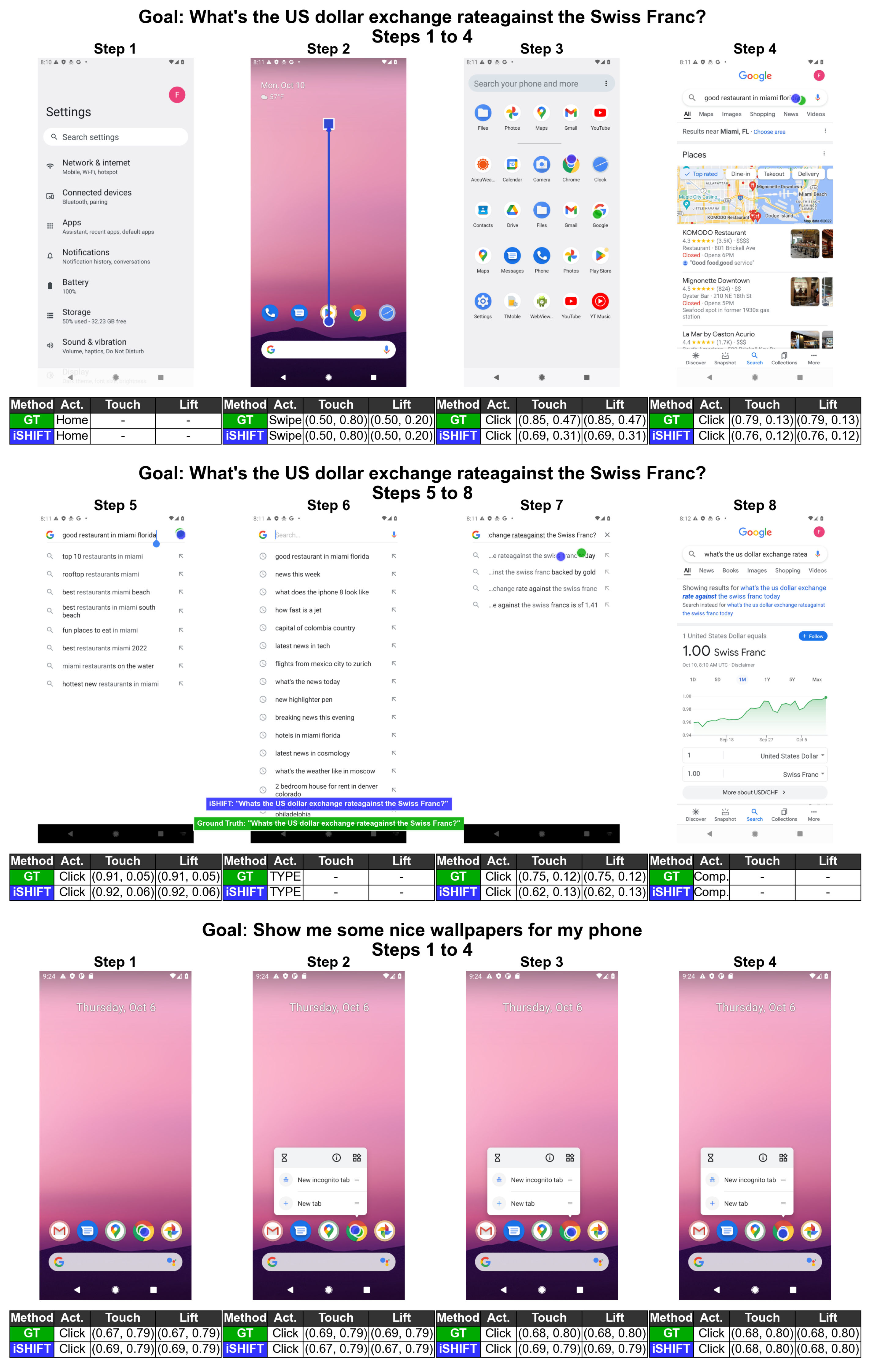}
    \vspace{-10pt}
    \caption{Episodes from the AITW dataset comparing \textcolor{blue}{iSHIFT} trajectories with the \textcolor{Green}{ground-truth} demonstrations. iSHIFT closely follows the intended interaction flow, and when deviations occur, the model selects alternative but valid action sequences that still achieve the correct task outcome.}
    \label{fig:episodes_7}
\end{figure*}

\begin{figure*}
    \centering
    \includegraphics[width=0.80\textwidth]{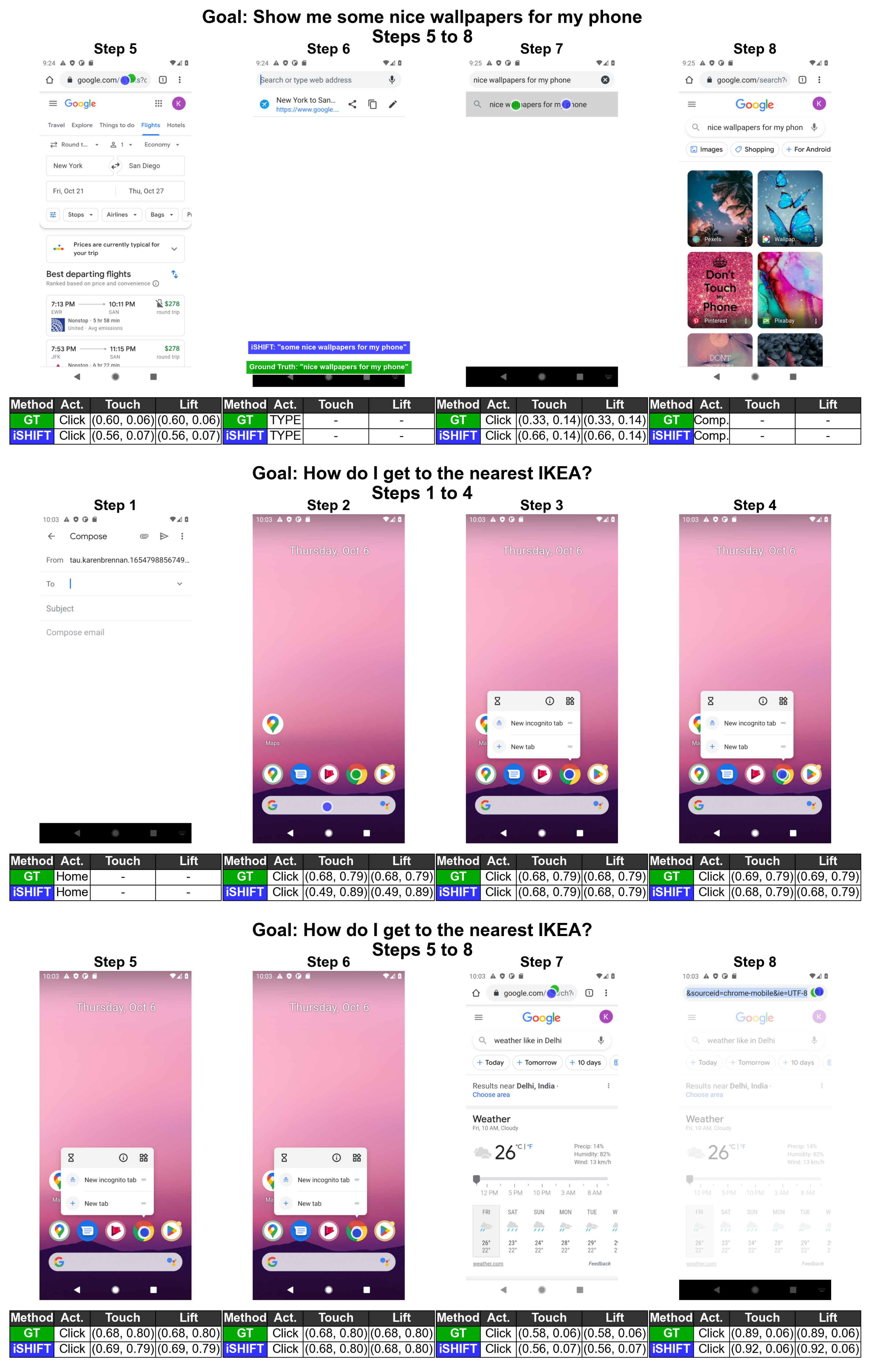}
    \vspace{-10pt}
    \caption{Episodes from the AITW dataset comparing \textcolor{blue}{iSHIFT} trajectories with the \textcolor{Green}{ground-truth} demonstrations. iSHIFT closely follows the intended interaction flow, and when deviations occur, the model selects alternative but valid action sequences that still achieve the correct task outcome.}
    \label{fig:episodes_8}
\end{figure*}

\begin{figure*}
    \centering
    \includegraphics[width=0.79\textwidth]{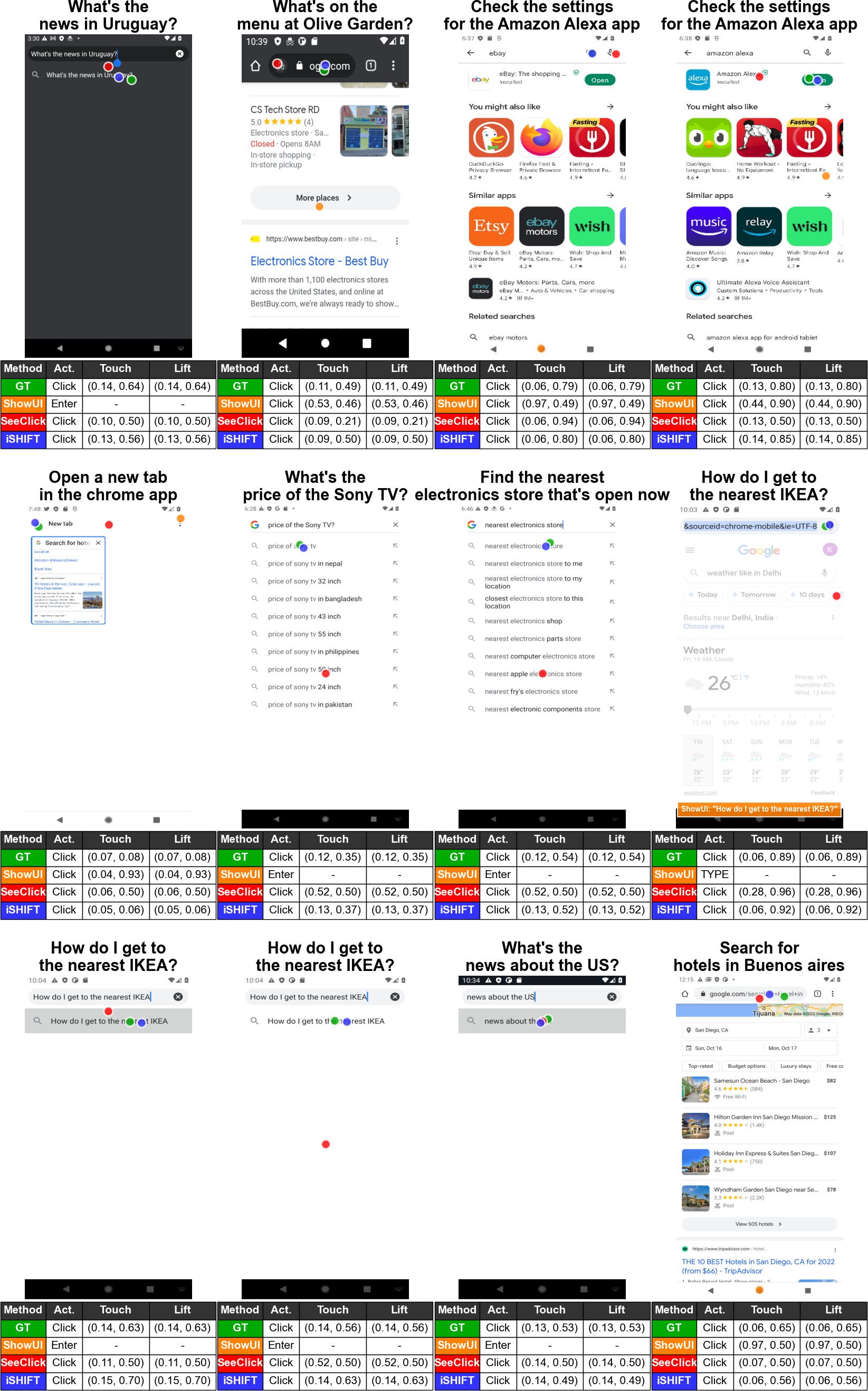}
    \caption{Qualitative comparison between \textcolor{blue}{iSHIFT}, \textcolor{orange}{ShowUI} and \textcolor{red}{SeeClick} on the AITW dataset. \textcolor{blue}{iSHIFT} consistently selects correct interaction steps and produces stable trajectories across diverse task scenarios, outperforming competing approaches.}
    \label{fig:comparisons_0}
\end{figure*}

\begin{figure*}
    \centering
    \includegraphics[width=0.79\textwidth]{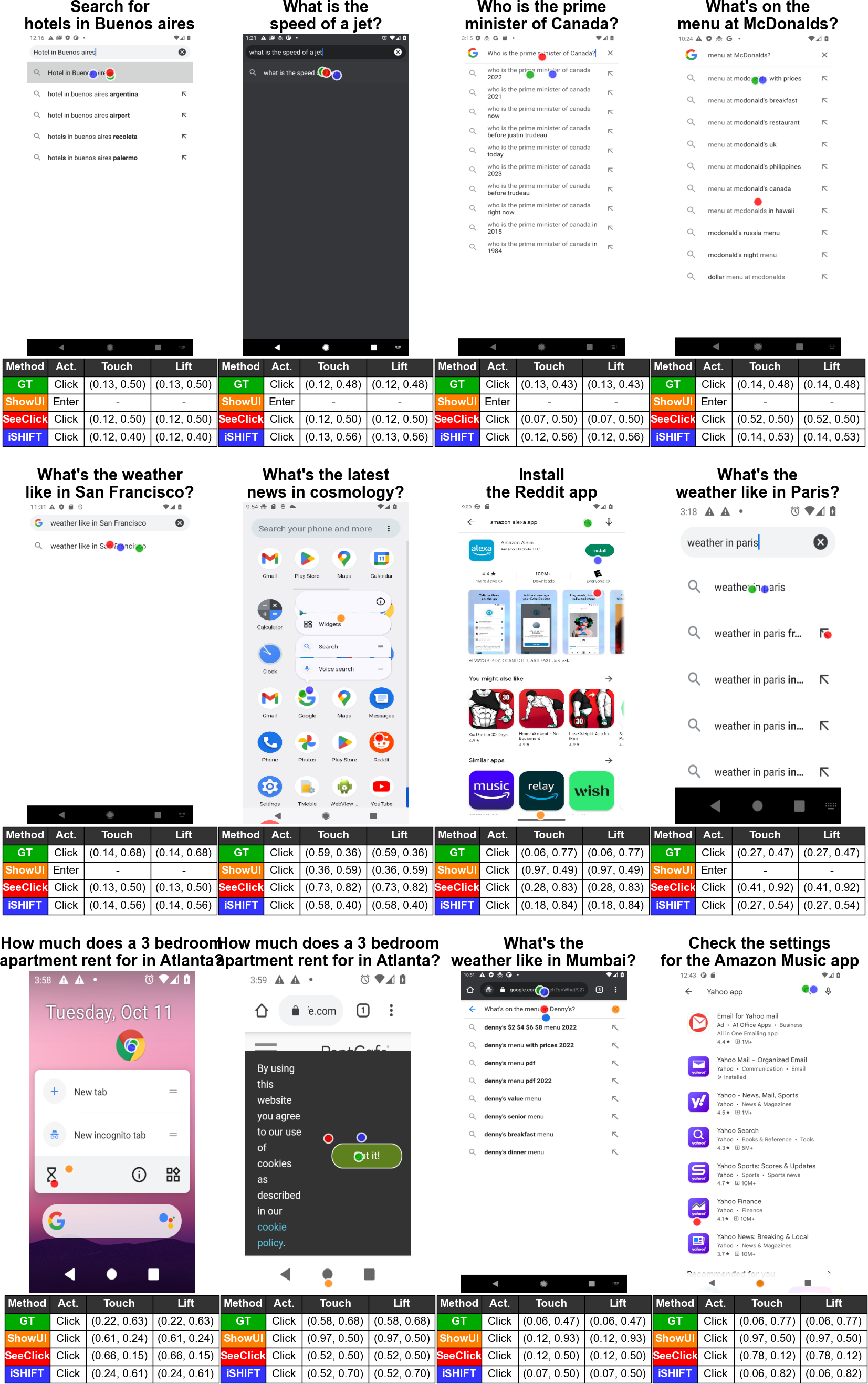}
    \caption{Qualitative comparison between \textcolor{blue}{iSHIFT}, \textcolor{orange}{ShowUI} and \textcolor{red}{SeeClick} on the AITW dataset. \textcolor{blue}{iSHIFT} consistently selects correct interaction steps and produces stable trajectories across diverse task scenarios, outperforming competing approaches.}
    \label{fig:comparisons_1}
\end{figure*}

\begin{figure*}
    \centering
    \includegraphics[width=0.79\textwidth]{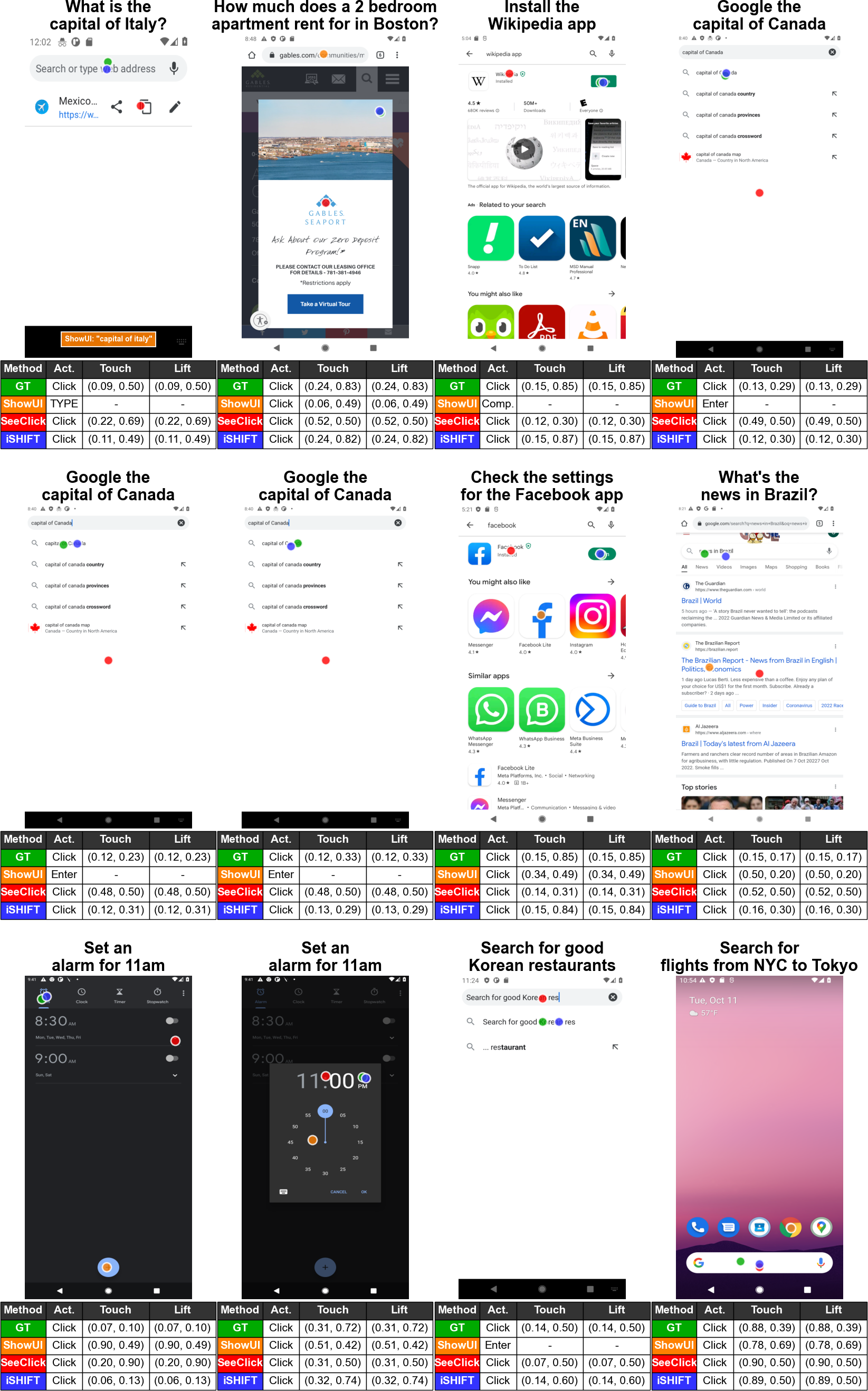}
    \caption{Qualitative comparison between \textcolor{blue}{iSHIFT}, \textcolor{orange}{ShowUI} and \textcolor{red}{SeeClick} on the AITW dataset. \textcolor{blue}{iSHIFT} consistently selects correct interaction steps and produces stable trajectories across diverse task scenarios, outperforming competing approaches.}
    \label{fig:comparisons_2}
\end{figure*}

\begin{figure*}
    \centering
    \includegraphics[width=\textwidth]{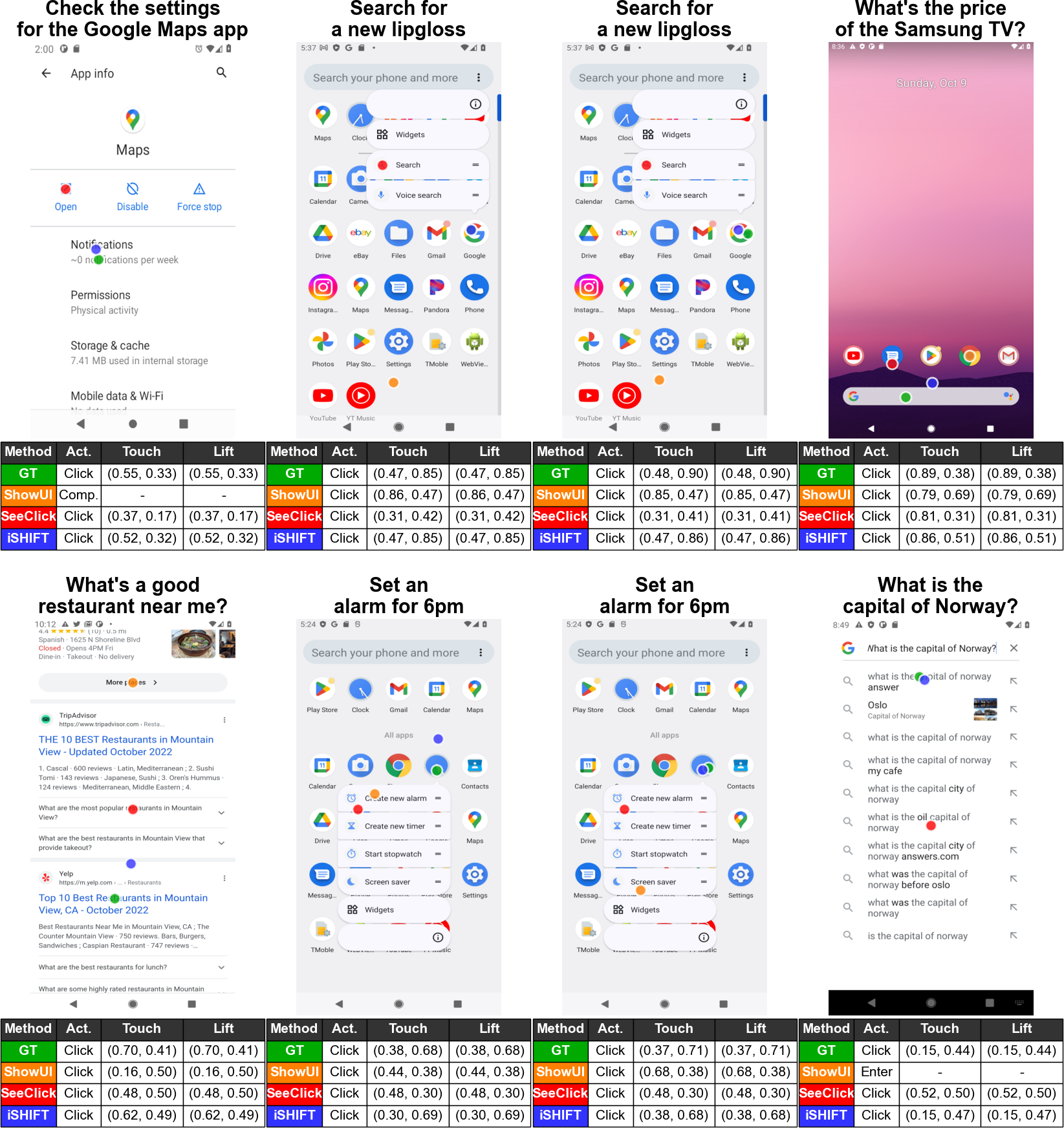}
    \caption{Qualitative comparison between \textcolor{blue}{iSHIFT}, \textcolor{orange}{ShowUI} and \textcolor{red}{SeeClick} on the AITW dataset. \textcolor{blue}{iSHIFT} consistently selects correct interaction steps and produces stable trajectories across diverse task scenarios, outperforming competing approaches.}
    \label{fig:comparisons_3}
\end{figure*}


\end{document}